\documentclass[3p,times,twocolumn]{elsarticle}
\usepackage{amsmath,amsfonts}
\usepackage{algorithmicx, algpseudocode}
\usepackage{float}
\usepackage{algorithm}
\usepackage{array}
\usepackage[caption=false,font=normalsize,labelfont=sf,textfont=sf]{subfig}
\usepackage{textcomp}
\usepackage{stfloats}
\usepackage{url}
\usepackage{verbatim}
\usepackage{graphicx}
\usepackage{caption}

\usepackage{multirow}
\usepackage{booktabs}
\usepackage{natbib}
\usepackage{tikz}
\usetikzlibrary{arrows.meta, positioning, fit, calc, shapes.geometric}
\usepackage{subcaption}
\usepackage{hyperref}
\makeatletter
\renewcommand{\corref}[1]{%
  \edef\cnotenum{\elsRef{#1}}%
  \edef\@corref{%
    \ifcase\cnotenum
    \or $\ast$%
    \or $\dagger$%
    \fi\hskip-1pt}}

\renewcommand{\cortext}[2][]{%
  \g@addto@macro\@cornotes{%
    \refstepcounter{cnote}\elsLabel{#1}%
    \def\thefootnote{%
      \ifcase\thecnote
      \or $\ast$%
      \or $\dagger$%
      \fi}%
    \footnotetext{#2}}}
\makeatother
\begin{document}



\title{HiBRIDGE: A Hierarchical Bayesian Neural Network Framework for Interpretable Dialogue Management in Group--Robot Interaction}

\author[1]{Massimiliano Nigro\corref{visit}}
\ead{massimiliano.nigro@polimi.it}
\author[2]{Hatice Gunes}
\ead{hg410@cam.ac.uk}
\author[1]{Micol Spitale\corref{equal}}
\ead{micol.spitale@polimi.it}
\author[2]{Fethiye Irmak Dogan\corref{equal}}
\ead{fid21@cam.ac.uk}

\affiliation[1]{organization={Department of Electronics, Information and Bioengineering, Politecnico di Milano},
            country={Italy}}
\affiliation[2]{organization={Affective Intelligence and Robotics (AFAR) Lab, Department of Computer Science and Technology, University of Cambridge},
            country={UK}}

\cortext[visit]{This research and the proposed technical methodology was initiated and completed in part while M. Nigro was a visiting PhD student at the AFAR Lab, Department of Computer Science and Technology, University of Cambridge.}
\cortext[equal]{Equal senior authorship.}




\begin{frontmatter}
    
\begin{abstract}
In multi-party human-robot interaction, a robot must continuously decide \emph{whom to address} and \emph{what to say} to participate effectively in the conversation. In real-world interactions, this is challenging because several behaviours may be plausible at the same time: a robot might continue a topic with one participant, involve another through a question, or address the whole group, with the appropriate choice depending on both whom it addresses and the interaction context. Current approaches remain limited in representing uncertainty when several behaviours are plausible and in structuring decisions into semantically meaningful intermediate steps that make robot decisions easier to interpret. Addressing these, we present \textbf{HiBRIDGE}, a hierarchical Bayesian neural network framework for group-robot dialogue management. Its Bayesian formulation enables uncertainty-aware prediction and robust learning from limited interaction data, while the hierarchical approach formulates behaviour selection as a structured, multi-stage decision process. We further use decision-tree surrogates to investigate whether this structure can support more interpretable explanations. Across three offline group-HRI datasets, our findings show that Bayesian formulations outperform their deterministic counterparts and several state-of-the-art baselines. Next, through an online study ($N=20$), we show that explanations derived from the hierarchical model are rated as more helpful for understanding robot behaviour and are preferred over those derived from the flat model. Finally, through our in-person study ($N=12$), we demonstrate the feasibility of HiBRIDGE for autonomous real-time group interaction, with both hierarchical and flat Bayesian variants positively perceived. Overall, HiBRIDGE combines strong predictive performance with a structured decision process that supports more interpretable explanations of robot behaviour. Our code will be publicly available at \href{https://github.com/Massimilianonigro/HiBridge/tree/main}{https://github.com/Massimilianonigro/HiBridge}.

\end{abstract}
\end{frontmatter}


\section{Introduction}

\begin{figure}[t]
    \centering
    \includegraphics[width=0.5\textwidth, trim=10 60 10 100, clip]{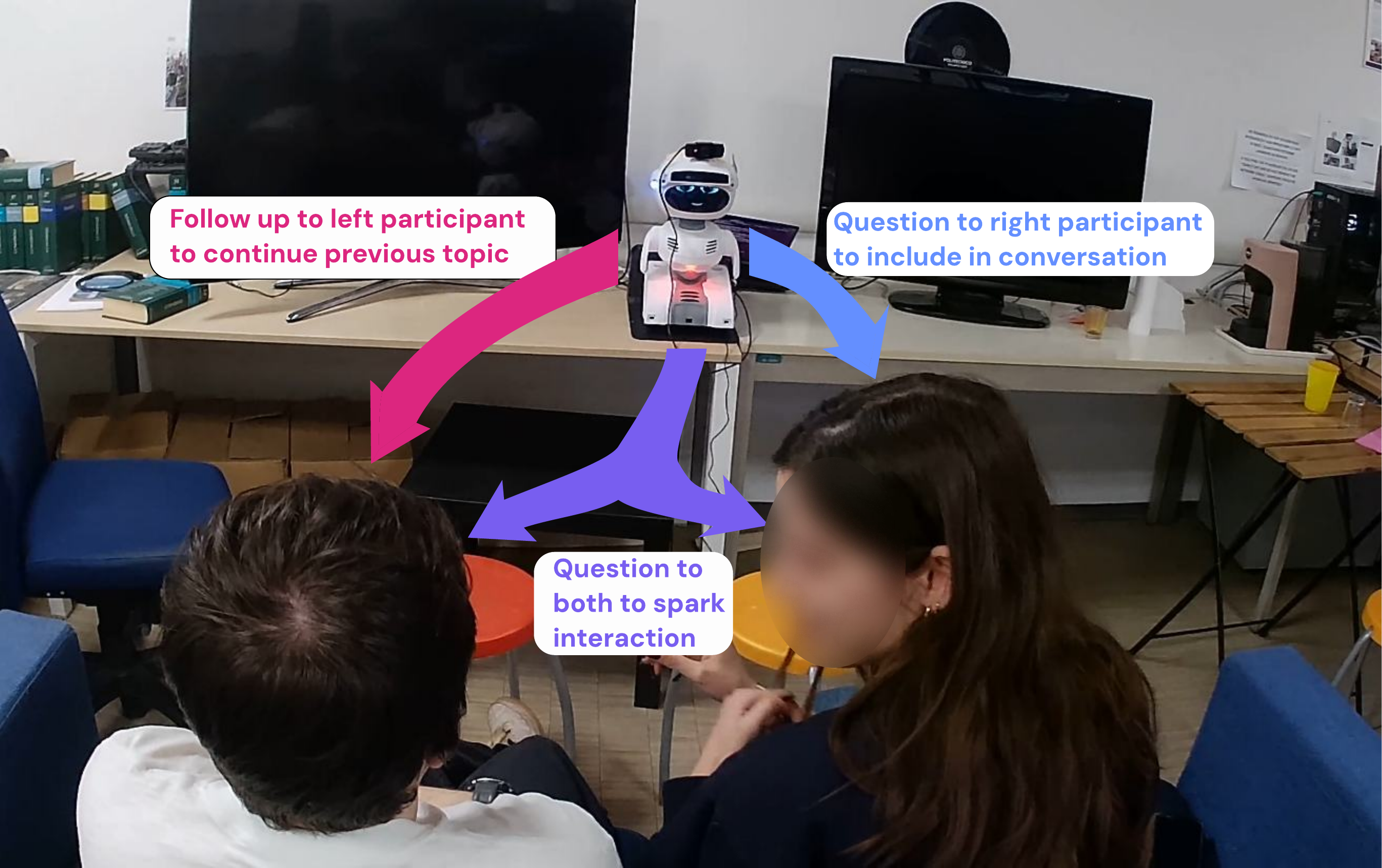}
    \captionsetup{justification=centering}
    \caption{Example of different plausible behaviours in a group conversation. Image taken from our in-person study.}
    \label{motivation}
\end{figure}

When interacting with a group, a robot must continuously decide both \textit{whom to address} and \textit{what to say} to support a smooth and appropriate interaction~\cite{nigro2025social, abbo2025fastmultipartyopenendedconversation, gillet2025templates}. In real-world multi-party conversations, however, several behaviours may be reasonable at the same moment. As illustrated in Figure~\ref{motivation}, the robot might continue a topic with one participant, involve another through a question, or address the whole group with a new topic. These choices are interdependent: what is appropriate to say depends partly on who is addressed, while the intended conversational content can in turn shape the choice of addressee. Appropriate behaviour also depends on the interaction context: a robot providing information in a clinic may primarily respond to user questions~\cite{addlesee2024multi}, whereas a robot moderating a discussion may need to intervene proactively and involve different participants~\cite{birmingham2020can, asadi2025not}. Group dialogue management therefore requires choosing among multiple plausible, interdependent, and context-sensitive behaviours \cite{engwall2021robot, grassi2025strategies}. This also makes transparency important: when a robot selects one behaviour over other reasonable alternatives, users should be able to understand the rationale for that choice and judge whether it was appropriate for the interaction.


Recent work on group-robot dialogue management has mainly followed two main directions: large language models (LLMs) and supervised learning approaches~\cite{abbo2025fastmultipartyopenendedconversation, gillet2025templates}. LLM-based systems can generate fluent, context-sensitive behaviour and have been successfully deployed in multi-party robotic interactions~\cite{addlesee2024multi}. However, when dialogue management is delegated directly to an LLM, controlling the robot's behaviour and explaining why a particular action was selected remain difficult. Supervised approaches instead can learn \textit{what to say} and \textit{whom to address} from interaction demonstrations, offering greater control over the resulting behaviour~\cite{gillet2025templates}. Prior work has modelled dependencies between these decisions through sequential prediction~\cite{gillet2025templates}, but does not explicitly structure the decision process into semantically meaningful intermediate levels that can support interpretation. In addition, supervised models are typically trained on small HRI datasets \cite{reimann2024survey}, making generalisation difficult, while deterministic formulations tend to be overconfident while uncertain in selecting among multiple plausible conversational behaviours~\cite{kristiadi2020being}. These limitations motivate models that can learn robustly from limited interaction data while providing a structured, probabilistic decision process whose intermediate stages can support explanations of robot behaviour.


In this work, we present \textbf{HiBRIDGE}, a hierarchical Bayesian neural network framework for deciding \textit{what to say} and \textit{whom to address} in multi-party human--robot interaction. The Bayesian formulation represents network parameters probabilistically, improving robustness when learning from limited HRI data. We further formulate behaviour selection as a hierarchical classification problem that decomposes the robot's decision into semantically meaningful stages, including whether its conversational act is initiative or responsive and whether it addresses an individual or the group. To capture the dependency between \textit{what to say} and \textit{whom to address}, we investigate both possible orderings: predicting the addressee before the dialogue act and predicting the dialogue act before the addressee. Finally, to assess whether this structured decision process supports interpretability, we fit decision-tree surrogate models~\cite{grassi2025strategies} to the model predictions and use their decision paths to generate explanations of the robot's behaviour.

Our work makes three main contributions. First, we introduce and evaluate HiBRIDGE across three group-HRI datasets, comparing Bayesian and deterministic formulations, their hierarchical and flat variants against LLM baselines, and a state-of-the-art supervised approach (\textbf{Contribution~1}). Our models outperform these state-of-the-art baselines, with the Bayesian formulation consistently improving predictive performance over its deterministic counterpart. Second, we investigate whether the intermediate structure exposed by the hierarchy can support more interpretable explanations of robot behaviour (\textbf{Contribution~2}). Using decision-tree surrogates to generate explanations from hierarchical and flat models, our online study shows that hierarchical explanations are perceived more favourably and are preferred over those produced by the flat model. Third, we integrate HiBRIDGE with an LLM-based response generator in a fully autonomous robotic system and evaluate it in an in-person group interaction study (\textbf{Contribution~3}). Both hierarchical and flat Bayesian models are positively perceived, demonstrating the feasibility of the framework for real-time group interaction. Overall, our results show that Bayesian modelling improves predictive performance, while hierarchical modelling provides a structured basis for more interpretable explanations without compromising predictive or interaction performance.

\section{Background and Related Works}

\subsection{Modeling Group-Robot Interaction}

For a robot to take part in an ongoing multi-party conversation, it must continuously decide \textit{whom to address} and \textit{what to say}~\cite{nigro2025social,abbo2025fastmultipartyopenendedconversation,gillet2025templates}. This group-robot dialogue management problem has largely been tackled from two directions in the literature: leveraging Large Language Models (LLMs) to select and generate conversational behaviour, or learning such behaviour directly from observed data through supervised models~\cite{gillet2025templates}.
One direction relies on LLMs, often combined with rule-based or perception components, to manage the robot's conversational behaviour~\cite{addlesee2024multi,abbo2025fastmultipartyopenendedconversation, zhu2025whom}. Addlesee et al.~\cite{addlesee2024multi} and Abbo et al.~\cite{abbo2025fastmultipartyopenendedconversation}, for example, incorporated LLMs into different architectures for multi-party interaction. Addlesee et al.~\cite{addlesee2024multi} developed an LLM-based dialogue system for a robot that answers patient questions in a memory clinic. Their approach determines whether the robot is being addressed in a multi-party conversation and, when appropriate, generates an LLM-based response. Abbo et al.~\cite{abbo2025fastmultipartyopenendedconversation} instead considered general open-ended conversations, integrating an LLM as a high-level dialogue manager alongside subsystems that provide contextual information. Given conversation transcripts, speaker labels, and the output of a turn-taking model indicating whether the robot should take or yield the turn, the LLM selects the robot's conversational behaviour. While these approaches offer substantial flexibility, relying on LLMs for behaviour selection makes the resulting behaviour more difficult to control and the rationale behind individual decisions harder to interpret~\cite{zhang2023large}.

An alternative is to formulate group-robot dialogue management as a supervised prediction problem, learning appropriate behaviour from observations of group interactions~\cite{gillet2025templates, 10731390, majlesi2023managing}. Gillet et al.~\cite{gillet2025templates}, for example, use Graph Neural Networks to learn \textit{whom to address} and \textit{what to say} while supporting groups of different sizes. They model the two decisions sequentially: a first Graph Neural Network predicts the addressee, and its output is then provided to a second network that predicts what the robot should say. Such approaches provide greater control over robot behaviour through the training observations, but are constrained by the small size of typical HRI datasets, which are costly to collect and can make supervised models susceptible to overfitting. Moreover, although these classifiers can achieve strong predictive performance, their decision process is not explicitly structured to support interpretation, making it difficult to explain how a particular robot behaviour was selected.

In our work, we build on the controllability of supervised learning while addressing these limitations through Bayesian neural networks and hierarchical classification. The Bayesian formulation supports robust learning from limited HRI data, while the hierarchical formulation exposes semantically meaningful intermediate stages of the decision process that can support interpretation. We further retain the flexibility of LLMs: we use the learned model to determine the dialogue act (\textit{what to say}) and addressee (\textit{whom to address}), while delegating only the generation of a contextually appropriate utterance to the LLM.

\subsection{Dialogue Acts Taxonomies}
\label{sec:dialogue_act_taxonomy}
Dialogue acts can be understood as the atomic units of a conversation, capturing the specific communicative intent behind an interaction \cite{bunt2005framework}. A dialogue act is typically characterised by two main components: its semantic content and its communicative function. The semantic content refers to the objects, propositions, or events that the act is about, while the communicative function specifies how the addressee should use this content to update their information state \cite{bunt2000dialogue, bunt2010towards}.
Several annotation schemes and taxonomies for dialogue acts have been proposed. One well-known example is Dialogue Act Markup in Several Layers (DAMSL) \cite{allen1997draft}, which represents dialogue acts along multiple dimensions: communicative status, indicating whether an utterance is intelligible and complete; information level, describing its semantic content; forward-looking function, capturing how the utterance constrains future beliefs and actions; and backwards-looking function, specifying how it relates to the preceding discourse.

Recently, the Machine Interaction Dialogue Act Scheme (MIDAS) \cite{yu2021midas} taxonomy has been developed specifically for human–machine interaction, with a particular emphasis on open-domain conversational settings. MIDAS is designed to support systems with limited capabilities for understanding their human interlocutors. It features a hierarchical structure, supports multi-label annotation, and distinguishes between two annotation layers: a semantic layer (e.g., questions, answers) and a functional layer (e.g., social conventions, incomplete requests). This taxonomy has been used in a variety of conversational systems for intent classification and utterance generation \cite{chi2022neural,fan2023athena,yu2019gunrock}.
In this work, we chose to focus on annotating specifically the semantic layer. As the hierarchical structure of MIDAS is particularly appealing for its expandability, we chose to adapt this taxonomy specifically for social robot interactions, combining relevant semantic elements from MIDAS with features from DAMSL to better fit our use case.
Dialogue acts allow us to formulate the \textit{what-to-say} problem in terms of high-level categories appropriate for social robot interactions.

\subsection{Bayesian Neural Network and Hierarchical Classification Approaches}
Bayesian Neural Networks (BNNs)~\cite{mullachery2018bayesian} are probabilistic extensions of traditional neural networks that incorporate uncertainty estimation through prior distributions over weights. Their uncertainty-aware structure allows them to be robust against overfitting and perform well on unseen data, even with small datasets~\cite{wilson2020bayesian, jospin2022hands}. This makes them particularly suited for Human-Robot Interaction (HRI) settings, where data collection is often limited by real-world deployment constraints and ethical considerations. In the HRI field, they have been used to learn spatial relations among objects in a human-robot collaboration task \cite{mccarthy2025bayesian} or to learn repulsive behaviours for safe hand avoidance in robotic arms \cite{shi2021bayesian}. They have also been applied to tasks such as learning spatial relationships in human–robot collaboration \cite{mccarthy2025bayesian} and modelling repulsive behaviours for safe hand avoidance in robotic arms \cite{shi2021bayesian}. In both cases, they demonstrated strong generalisation performance and robustness when applied to new, unseen scenarios.

Hierarchical classification methods \cite{silla2011survey} address the challenge of assigning instances to labels organised within a predefined taxonomy or hierarchy, allowing predictions at multiple levels of abstraction. 
A key advantage of these methods is their ability to capture and exploit correlations among classes at different hierarchical levels, in contrast to traditional classification (named flat) approaches that treat class labels as independent and unrelated~\cite{silla2011survey,gordon11996hierarchical}. \textbf{\emph{To the best of our knowledge, our work is the first to apply hierarchical classification in the Human–Robot Interaction domain, leveraging its multi-level structure to expose intermediate stages of the robot’s decision process, thereby improving transparency.}} 
In our paper, we investigate whether combining Bayesian Neural Networks with hierarchical classification can improve interpretability and predictive performance in our setting, where data is both limited and severely imbalanced.

\section{Problem Formulation}
\begin{figure*}[t]
    \centering
    \includegraphics[width=\textwidth]{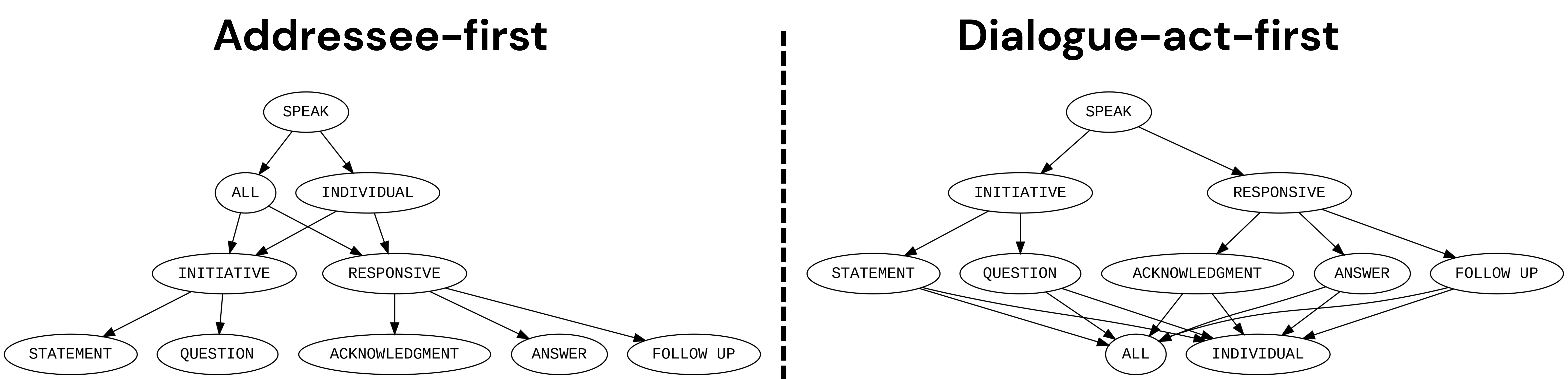}
    \caption{Representation of addressee-first and dialogue-act-first hierarchies.}
    \label{fig:hierarchy_image}
\end{figure*}

\subsection{Task Definition}
\label{definition}

We formalise the problem of determining both what dialogue act to perform and whom to address as a hierarchical classification task over a composite hierarchy. At each conversational step $t$, given the conversational state $\mathbf{x}_t$, the robot first decides whether to speak or remain silent. We denote this decision as $s_t \in \{0,1\}$, where $s_t=0$ corresponds to remaining silent and $s_t=1$ proceeds to conversational behaviour selection. When $s_t=1$, the robot predicts a complete conversational behaviour:
\begin{equation}
    y_t=(d_t,a_t),
    \qquad
    d_t\in\mathcal{D},
    \quad
    a_t\in\mathcal{A},
\end{equation}%
where $\mathcal{D}$ denotes the set of dialogue-act classes and $\mathcal{A}$ the set of addressee classes. The composite hierarchy contains one component dedicated to the choice of the \textbf{addressee} and another dedicated to the choice of the \textbf{dialogue act}.

We define two alternative hierarchical structures, corresponding to the two possible orderings of these decisions:
\begin{equation}
    \mathcal{H}\in
    \left\{
        \mathcal{H}_{A\rightarrow D},
        \mathcal{H}_{D\rightarrow A}
    \right\},
\end{equation}
where $\mathcal{H}_{A\rightarrow D}$ prioritises addressee selection followed by dialogue-act choice (\emph{addressee-first}), while $\mathcal{H}_{D\rightarrow A}$ reverses this ordering (\emph{dialogue-act-first}) -- see Figure~\ref{fig:hierarchy_image}. In both cases, a complete behaviour corresponds to a unique root-to-leaf path:
\begin{equation}
    \pi_t=(v_0, v_1,v_2,\ldots,v_K)
    \in \Pi(\mathcal{H}),
\end{equation}
where $\Pi(\mathcal{H})$ denotes the set of valid paths through the hierarchy. The classification objective is therefore to identify the path $\hat{\pi}_t$ corresponding to the robot's complete decision.

The \emph{addressee} component of the hierarchy distinguishes between addressing the entire group and addressing an individual. Although the general formulation may further resolve the individual branch into specific participants, in the streamlined hierarchy used in this work we restrict the addressee space to
    $\mathcal{A}
    =
    \{
        \textsc{All},
        \textsc{Individual}
    \}$.

The \emph{dialogue-act} component follows the semantic taxonomy detailed in the next section. Given the limited size of human--robot interaction datasets, we adopt a streamlined version of the hierarchy to mitigate data imbalance and reduce the number of unique classes. The first dialogue-act level distinguishes between initiative and responsive behaviour while the second level predicts the corresponding dialogue-act category:

\begin{gather}
    d_t^{(1)}
    \in
    \mathcal{D}^{(1)}
    =
    \{
        \textsc{Initiative},
        \textsc{Responsive}
    \},\\
    d_t^{(2)}
    \in
    \begin{cases}
        \{
        \textsc{Question},
        \textsc{Statement}
        \},
        & d_t^{(1)}=\textsc{Initiative},
        \\[2mm]
        \left\{
        \begin{aligned}
            &\textsc{Acknowledgement},\\
            &\textsc{Follow\mbox{-}up},
            \textsc{Answer}
        \end{aligned}
        \right\},
        & d_t^{(1)}=\textsc{Responsive},
    \end{cases}
\end{gather}%
$d_t^{(2)}$ corresponds to the final dialogue-act prediction $d_t$. The addressee and dialogue-act hierarchies are combined by attaching the leaf nodes of the first component to the root of the second component and subsequently pruning the redundant root node. As a result, the two possible decision paths can be summarised as follows:
\begin{equation}
    \pi_t =
    \begin{cases}
        \textsc{Speak}
        \rightarrow
        a_t
        \rightarrow
        d_t^{(1)}
        \rightarrow
        d_t^{(2)},
        & \text{addressee-first},
        \\[2mm]
        \textsc{Speak}
        \rightarrow
        d_t^{(1)}
        \rightarrow
        d_t^{(2)}
        \rightarrow
        a_t,
        & \text{dialogue-act-first}.
    \end{cases}
\end{equation}

The \textsc{Speak} root is fixed conditional on $s_t=1$, and the subsequent nodes constitute the behaviour-selection decisions. Therefore, together with the preliminary speak-or-silence decision, the complete task can be represented as
$    \mathbf{x}_t
    \rightarrow
    s_t
    \rightarrow
    \pi_t
    \rightarrow
    (d_t,a_t),$
where hierarchical behaviour prediction is performed only when $s_t=1$, when the robot needs to speak. These structures represent the task-specific instantiations of the generic hierarchy used by HiBRIDGE, defined formally in Section~\ref{sec:hibridge}.

\subsection{Dialog Act Taxonomy}
Our dialogue act taxonomy preserves the core semantic structure of MIDAS~\cite{yu2021midas}, particularly the distinction between initiative and responsive acts, while introducing targeted modifications to better suit social, multi-party human–robot interaction. These adjustments are largely inspired by DAMSL~\cite{allen1997draft} and by analysis of the three multi-party human–robot interaction datasets used in this work. Our taxonomy includes \textbf{initiative} and \textbf{responsive} acts.

First, \emph{initiative acts} begin new conversation threads: in our taxonomy, they consist of Questions and Statements. Questions include factual questions (seeking specific information, e.g., “How many landers does Germany have?”), opinion questions (eliciting subjective responses, e.g., “How was your time at the mall today?”), offers (e.g., “Would you like to try a quiz?”), and small talk (e.g., “How are you?”). Statements introduce new topics or fulfil social conventions through declarative contributions (e.g., “Hello there” or “I see you have bags with you”).
To formulate our taxonomy, starting from MIDAS, we removed the command category, as it was not representative of the multi-party interaction settings we examine. To maintain adequate coverage, we introduce a broader Statement category, inspired by DAMSL, to capture declarative contributions that initiate or shift conversational topics. We also extend the Question category to subsume functions previously covered by commands, incorporating Offer (following DAMSL) and adding Small Talk to capture socially motivated, non-information-seeking utterances.

Next, \emph{responsive acts} continue existing conversational threads: in our taxonomy, they are divided into Acknowledgements, Follow-ups, and Answers. Acknowledgements address prior contributions through brief comments that affirm without elaboration (e.g., “That’s great”) or opinions that add evaluative content (e.g., “Oh, that’s lovely”). Follow-ups extend the dialogue by posing related questions (e.g., “Is that your final answer?”). Answers respond directly to prior questions and are categorised as positive, negative, or other. 
In our taxonomy, we group MIDAS Opinions and non-opinionated Statements (renamed comments) into a unified Acknowledgement category, as both serve similar functions in signalling understanding and grounding. We also introduce a Follow-up category to reflect patterns in our data where participants and robots extend prior turns through clarifying or elaborative questions.

Overall, the changes in initiative and responsive acts preserve the strengths of MIDAS while improving its ability to capture the semantic and interactional patterns observed in our data.

\section{HiBRIDGE Framework}
\label{sec:hibridge}

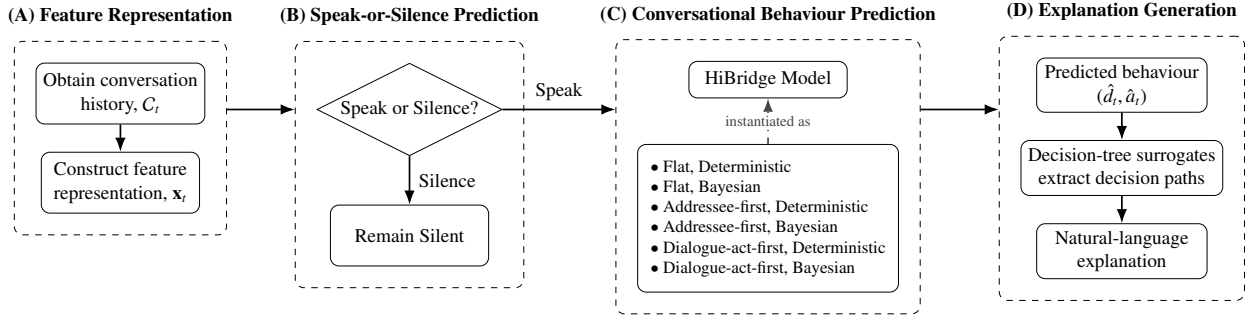
\begin{figure*}[t]
\centering
\resizebox{\textwidth}{!}{%
\begin{tikzpicture}[
    >=Latex,
    every node/.style={font=\small},
    block/.style={
        draw,
        rounded corners,
        align=center,
        minimum height=1cm,
        minimum width=2.6cm
    },
    smallblock/.style={
        draw,
        rounded corners,
        align=center,
        minimum height=0.6cm,
        minimum width=2.6cm,
        font=\small
    },
    explanationblock/.style={
        draw,
        rounded corners,
        align=center,
        minimum height=0.75cm,
        minimum width=2.8cm,
        font=\small
    },
    decision/.style={
        diamond,
        draw,
        align=center,
        aspect=2,
        inner sep=1pt,
        minimum width=2.2cm
    },
    groupbox/.style={
        draw,
        dashed,
        rounded corners,
        inner sep=0.35cm
    },
    line/.style={
        draw,
        -Latex,
        thick
    },
    arrowlabel/.style={
        fill=white,
        inner sep=2pt,
        font=\small
    },
    selectline/.style={
        draw,
        dashed,
        -{Latex[length=2.2mm]},
        thick,
        gray!55!black
    }
]
\node[block] (history) at (0,0.55cm)
{Obtain conversation\\
history, $\mathcal{C}_t$};
\node[block, below=0.45cm of history] (features)
{Construct feature\\
representation, $\mathbf{x}_t$};
\node[
    groupbox,
    fit=(history)(features)
] (featuregroup) {};
\draw[line]
(history.south) --
(features.north);
\coordinate (speakcenter)
at ($(featuregroup.east)+(3cm,-0.8cm)$);
\node[decision] (speak)
at ($(speakcenter)+(0,1.27cm)$)
{Speak or Silence?};
\node[block] (silent)
at ($(speakcenter)+(0,-0.80cm)$)
{Remain Silent};
\node[
    groupbox,
    fit=(speak)(silent)
] (speakcheck) {};
\coordinate (flowrow) at (speak);
\node[smallblock, anchor=west, xshift=2.7cm,  yshift=0.49cm] (behaviour)
at (speakcheck.east |- flowrow)
{HiBridge Model};
\node[
    draw,
    rounded corners,
    align=left,
    font=\footnotesize,
    below=0.75cm of behaviour,
    inner sep=6pt
] (legend)
{%
\begin{tabular}[t]{@{}l@{}}
$\bullet$ Flat, Deterministic\\
$\bullet$ Flat, Bayesian\\
$\bullet$ Addressee-first, Deterministic\\
$\bullet$ Addressee-first, Bayesian\\
$\bullet$ Dialogue-act-first, Deterministic\\
$\bullet$ Dialogue-act-first, Bayesian
\end{tabular}%
};
\draw[selectline]
(legend.north) --
node[arrowlabel, font=\scriptsize, inner sep=1.5pt] {instantiated as}
(behaviour.south);
\node[
    groupbox,
    fit=(behaviour)(legend)
] (models) {};
\node[
    explanationblock,
    anchor=west,
    xshift=1.9cm,
    yshift=0.4cm
] (pred)
at (models.east |- flowrow)
{Predicted behaviour\\
$(\hat{d}_t,\hat{a}_t)$};
\node[
    explanationblock,
    below=0.45cm of pred
] (surrogate)
{Decision-tree surrogates\\
extract decision paths};
\node[
    explanationblock,
    below=0.45cm of surrogate
] (explanation)
{Natural-language\\
explanation};
\node[
    groupbox,
    fit=(pred)(surrogate)(explanation)
] (explanationgroup) {};
\node[font=\small, anchor=south] at ($(featuregroup.north)+(0,4pt)$)
{\textbf{(A) Feature Representation}};
\node[font=\small, anchor=south] at ($(speakcheck.north)+(0,4pt)$)
{\textbf{(B) Speak-or-Silence Prediction}};
\node[font=\small, anchor=south] at ($(models.north)+(0,4pt)$)
{\textbf{(C) Conversational Behaviour Prediction}};
\node[font=\small, anchor=south] at ($(explanationgroup.north)+(0,4pt)$)
{\textbf{(D) Explanation Generation}};
\draw[line]
(featuregroup.east |- flowrow) --
(speakcheck.west |- flowrow);
\draw[line]
(speak.south) --
node[arrowlabel, right=2pt] {Silence}
(silent.north);
\draw[line]
(speak.east |- flowrow) --
node[arrowlabel, above=2pt] {Speak}
(models.west |- flowrow);
\draw[line]
(models.east |- flowrow) --
(explanationgroup.west |- flowrow);
\draw[line]
(pred.south) --
(surrogate.north);
\draw[line]
(surrogate.south) --
(explanation.north);
\end{tikzpicture}%
}
\caption{Overall HiBRIDGE pipeline. The system first encodes the current conversation $\mathcal{C}_t$ into a feature representation $\mathbf{x}_t$ and determines whether the robot should speak. If the robot remains silent, no further behaviour is selected. Otherwise, the HiBridge model predicts the dialogue act and addressee, $(\hat{d}_t,\hat{a}_t)$; HiBridge is instantiated as one of six trained variants, combining a behaviour ordering (flat, addressee-first, or dialogue-act-first) with an inference strategy (deterministic or Bayesian), as listed in the legend. Decision-tree surrogates subsequently extract the corresponding decision paths from predicted behaviour, which are then used to generate a natural-language explanation.}
\label{fig:hibridge_pipeline}
\end{figure*}

HiBRIDGE models group-robot dialogue management as a mapping from the current conversational state to a robot behaviour. At each decision step $t$, the input is a feature vector $\mathbf{x}_t$ describing the recent interaction. The system first determines whether the robot should speak (Section~\ref{slience_model}) and, when it does, outputs a joint prediction of the dialogue act $d_t$ (\textit{what to say}) and addressee $a_t$ (\textit{whom to address}). For the subsequent behaviour-prediction stage, we consider both \textbf{flat} and \textbf{hierarchical} architectures (Section~\ref{hierarhy model}), and instantiate each architecture using \textbf{deterministic} and \textbf{Bayesian} neural networks (Section~\ref{Baysian_model}). We further use decision-tree surrogate models to extract interpretable decision paths from the learned models (Section~\ref{explanation_model}). The overall HiBRIDGE pipeline can be seen in Figure~\ref{fig:hibridge_pipeline}. While Section~\ref{definition} specifies the particular dialogue-act and addressee hierarchy used in this work, the HiBRIDGE formulation below is agnostic to the semantics and depth of the hierarchy. 

\subsection{Feature Representation}
\label{sec:feature_representation}
At each decision step $t$, the feature vector $\mathbf{x}_t$ is built by combining information from two time horizons. A sliding window keeps the six most recent conversational turns; for each turn in the window, we record who spoke, who was addressed, the dialogue act performed, the utterance's duration, and the length of any preceding silence. As the conversation progresses, this window is updated incrementally: the oldest turn is dropped, and the newest is added, so $\mathbf{x}_t$ always reflects the most recent local context. Positions in the window with no corresponding turn (e.g. at the start of a conversation) are padded with a dedicated placeholder value.

In parallel, we maintain a set of interaction-level statistics that summarise the conversation as a whole, rather than any single turn. We represent the group as a directed graph in which each node is a group member (including the robot) and each edge weight is the cumulative time that member has spent speaking directly to another. From this graph, we derive, for every group member, the total time they have spent addressing each other member (the graph's edges), the total time they have been addressed by anyone (in-degree), and the total time they have spent speaking to anyone (out-degree). Unlike the turn-level window, these statistics are cumulative: they are updated after every turn and never discarded, so they capture patterns of participation over the whole interaction rather than only its recent history.

All features are computed assuming a maximum group size of four (three people and a robot), the largest group size across our datasets; conversations with fewer participants have the unused positions padded. Categorical features (speaker, addressee, dialogue act) are one-hot encoded, while numerical features (durations, silences, and the interaction-level statistics) are standardised using the mean and standard deviation of each feature within its dataset. More details on the resulting composition of the final 158-dimensional representation are in the appendix section ~\ref{app:feature_set}.

\subsection{Speak-or-Silence Prediction}
\label{slience_model}

To mitigate the class imbalance that arises when comparing every instance in which the robot remains silent against each individual addressee/dialogue-act combination, we decompose the decision process into two stages, beginning with a speak/no-speak decision. Before selecting a dialogue act and addressee, the system determines whether the robot should take a conversational turn.
We model this as a separate binary classification problem, $s_t \in \{0,1\}$, where $s_t=1$ indicates that the robot should speak and $s_t=0$ that it should remain silent, using a feedforward network:
\begin{equation}
    p(s_t=1\mid\mathbf{x}_t)
    =
    \sigma
    \left(
    f_{\mathrm{speak}}(\mathbf{x}_t)
    \right),
\end{equation}%
where $\sigma(\cdot)$ denotes the sigmoid function. The network is trained using binary cross-entropy:
\begin{equation}
    \mathcal{L}_{\mathrm{BCE}}
    =
    -
    \sum_{i=1}^{N}
    \left[
        s_i\log p_i
        +(1-s_i)\log(1-p_i)
    \right].
\end{equation}%
If $\hat{s}_t=1$, the system enters the conversational behaviour-selection hierarchy at the \textsc{Speak} root and proceeds to predict the dialogue act and addressee.

\subsection{Conversational Behaviour Prediction}
Conditioned on the robot deciding to speak ($\hat{s}_t=1$), conversational behaviour prediction proceeds from the \textsc{Speak} root. 
Let $\mathcal{D}$ denote the set of dialogue-act classes and $\mathcal{A}$ the set of addressee classes. The complete robot behaviour is represented as:
\begin{equation}
    y_t = (d_t,a_t), \qquad
    d_t \in \mathcal{D}, \quad a_t \in \mathcal{A},
\end{equation}%
where $d_t$ denotes the final dialogue-act outcome, abstracting over any intermediate dialogue-act levels defined by the task-specific hierarchy. We investigate three architectures for estimating $p(y_t \mid \mathbf{x}_t)$: a flat model, an addressee-first hierarchical model, and a dialogue-act-first hierarchical model. Each architecture is implemented in both deterministic and Bayesian form, resulting in six model variants. 

\begin{figure*}[t]
    \centering
    \includegraphics[width=0.9\textwidth]{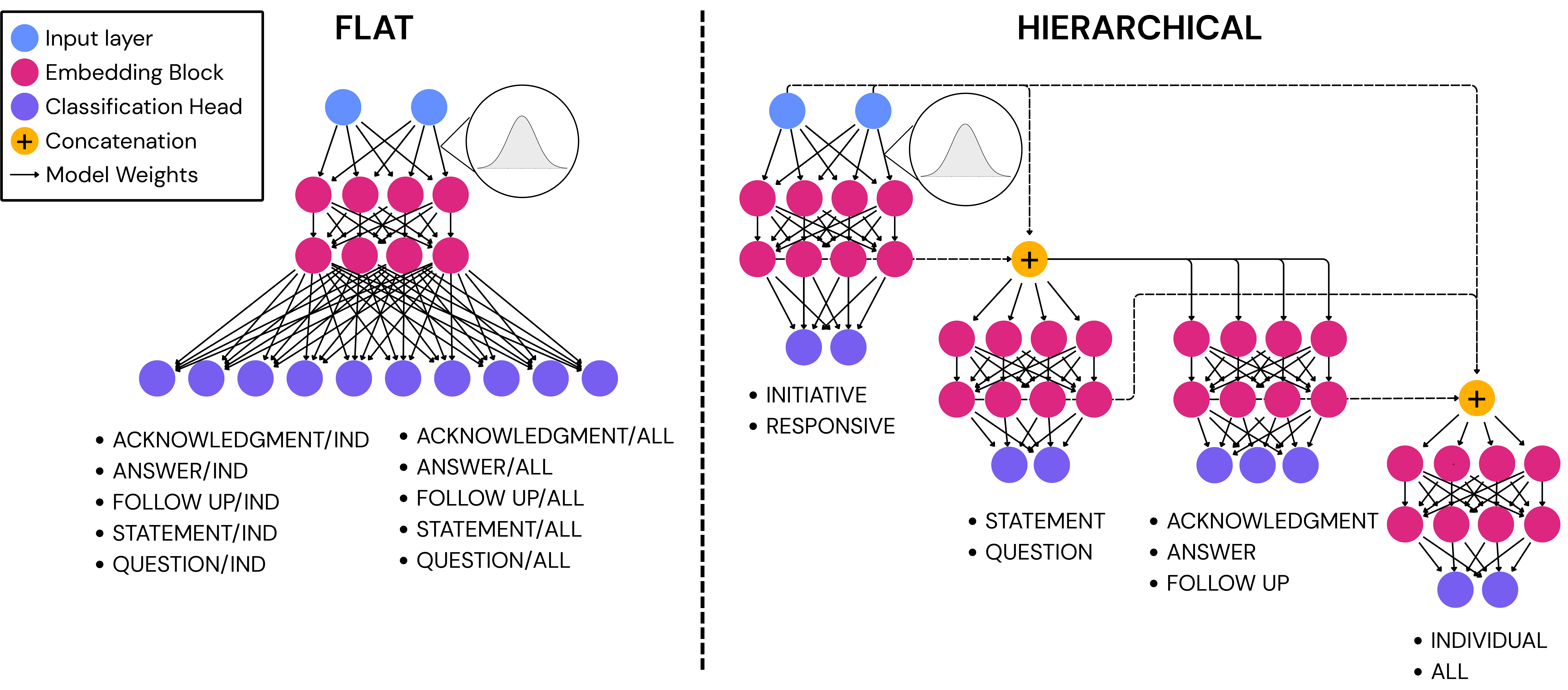}
    \captionsetup{justification=centering}
    \caption{Flat and hierarchical (dialogue-act-first) architectures for HiBRIDGE conversational behaviour prediction. Each model weight is drawn from a learned gaussian distribution. In the flat architecture (left), a single classification head predicts one of the ten joint dialogue-act/addressee classes directly. In the hierarchical architecture (right), each classification head instead predicts one level of the hierarchy (initiative/responsive, then dialogue-act category, then addressee), with its embedding concatenated back with the original input at each subsequent level.}
    \label{sturture}
\end{figure*}

\subsubsection{Flat and Hierarchical Architectures}
\label{hierarhy model}
We first describe flat and hierarchical neural network architectures used for predicting the dialogue act and addressee selection.

\textbf{Flat Architecture:} The flat architecture treats each valid dialogue-act--addressee combination as a single class. Let $\mathcal{Y} \subseteq \mathcal{D}\times\mathcal{A}$ denote the set of valid joint behaviours. A feedforward neural network first maps the conversational features to a latent representation:
\begin{gather}
    \mathbf{h}^{(0)} = \mathbf{x}_t,\\
    \mathbf{h}^{(l)}
    =
    \phi^{(l)}
    \left(
        \mathbf{W}^{(l)}\mathbf{h}^{(l-1)}
        +
        \mathbf{b}^{(l)}
    \right),
    \qquad l=1,\ldots,L.
\end{gather}%
where each hidden layer is followed by dropout and batch normalisation. The logits produced by the final layer, the probability of each joint behaviour and the selected behaviour are obtained as:
\begin{gather}
    \mathbf{z}
    =
    \mathbf{W}^{(o)}\mathbf{h}^{(L)}
    +
    \mathbf{b}^{(o)},\\
    p(y_t=c\mid\mathbf{x}_t)
    =
    \frac{\exp(z_c)}
    {\sum_{c'\in\mathcal{Y}}\exp(z_{c'})},\\
    \hat{y}_t
    =
    \arg\max_{c\in\mathcal{Y}}
    p(y_t=c\mid\mathbf{x}_t).
\end{gather}

In this formulation, the dialogue act and addressee are predicted jointly without exposing intermediate decisions.

\textbf{Hierarchical Architecture:} 
The hierarchical formulation instead represents a robot behaviour as a path through a hierarchy. Let $\mathcal{H}=(\mathcal{V},\mathcal{E})$ denote the hierarchy, where $\mathcal{V}$ contains decision nodes and $\mathcal{E}$ defines parent--child relationships. The nodes in $\mathcal{V}$ may represent any sequence of intermediate decisions; in our task, they correspond to the addressee and dialogue-act decisions defined in Section~\ref{definition}. Conditioned on $\hat{s}_t=1$, let $v_0=\textsc{Speak}$ denote the root node of the hierarchy. A complete behaviour from the root to a leaf node corresponds to a path as described before:
$    \pi =
    (v_0, v_1,v_2,\ldots,v_K)$, where $v_1,\ldots,v_K$ denote the successive decisions made
by the behaviour-selection hierarchy. 

At each hierarchical level $k$, HiBRIDGE contains an embedding block and classification heads (see Figure~\ref{sturture}). The embedding at level $k$ is computed from the original conversational representation and the embedding inherited from the parent level:
\begin{equation}
    \mathbf{h}_k
    =
    f_k
    \left(
    [\mathbf{x}_t;\mathbf{h}_{\mathrm{pa}(k)}]
    \right),
\end{equation}

where $[\cdot;\cdot]$ denotes concatenation and $f_k(\cdot)$ is a feedforward embedding block composed of hidden layers, dropout, and batch normalisation. For a parent node $v$, the corresponding classification head produces a conditional distribution over its children:
\begin{equation}
    p(u\mid v,\mathbf{x}_t)
    =
    \operatorname{softmax}
    \left(
    g_v(\mathbf{h}_k)
    \right)_u,
    \qquad
    u\in\operatorname{ch}(v),
\end{equation}%
where $\operatorname{ch}(v)$ denotes the children of node $v$. The probability of a complete behaviour is obtained from the conditional decisions along its path:
\begin{equation}
    \log p(\pi\mid\mathbf{x}_t)
    =
    \sum_{k=1}^{K}
    \log
    p(v_k \mid v_{k-1},\mathbf{x}_t).
\end{equation}%
Finally, the predicted behaviour corresponds to the valid path with maximum probability:
\begin{equation}
    \hat{\pi}_t
    =
    \arg\max_{\pi\in\Pi(\mathcal{H})}
    p(\pi\mid\mathbf{x}_t).
\end{equation}%
where $\Pi(\mathcal{H})$ denotes the set of valid root-to-leaf paths. We consider two orderings of the hierarchy:
\begin{equation}
p(d_t,a_t\mid\mathbf{x}_t)
=
\begin{cases}
p(a_t\mid\mathbf{x}_t)\,
p(d_t\mid a_t,\mathbf{x}_t),
& \text{addressee-first},\\[2mm]
p(d_t\mid\mathbf{x}_t)\,
p(a_t\mid d_t,\mathbf{x}_t),
& \text{dialogue-act-first}.
\end{cases}
\end{equation}

The dialogue-act component is itself hierarchical. It first distinguishes between \textit{initiative} and \textit{responsive} behaviour before predicting the corresponding dialogue-act category. The addressee component distinguishes whether the robot addresses an individual or the entire group. Consequently, unlike the flat formulation, the hierarchical architecture exposes the intermediate decisions leading to the final behaviour.

\subsubsection{Deterministic and Bayesian Formulations}
\label{Baysian_model}
The architectures described above are instantiated using both deterministic and Bayesian neural networks. 

\textbf{Deterministic Formulation:} 
In the deterministic formulation, the network parameters are represented by fixed values learned during training, and the standard classification objective is cross-entropy:
\begin{gather}
    \boldsymbol{\theta}
    =
    \{\mathbf{W}^{(l)},\mathbf{b}^{(l)}\}_{l=1}^{L},\\
    \mathcal{L}_{\mathrm{CE}}
    =
    -\sum_{i=1}^{N}
    \log
    p_{\boldsymbol{\theta}}
    (y_i\mid\mathbf{x}_i).
\end{gather}

We additionally evaluate class-weighted cross-entropy to account for the strong imbalance in dialogue-act--addressee combinations. For hierarchical models, we also consider a hierarchical penalty that takes into account the distance between the predicted and target classes in the hierarchy. Let $ d_{\mathcal{H}}(y_i,\hat{y}_i)$ denote this distance; the training objective can be written as follows:
\begin{equation}
    \mathcal{L}_{\mathrm{det}}
    =
    \mathcal{L}_{\mathrm{CE}}
    +
    \lambda_{\mathcal{H}}
    \mathcal{L}_{\mathcal{H}},
\end{equation}%
where $\lambda_{\mathcal{H}}$ controls the contribution of the hierarchical penalty. This assigns smaller penalties to errors between nearby classes in the hierarchy than to errors across more distant branches.

\textbf{Bayesian Formulation:} 
In the Bayesian formulation, network parameters are represented probabilistically. Rather than learning a single parameter estimate $\boldsymbol{\theta}$, we place distributions over the network weights,
    $\boldsymbol{\theta}
    \sim
    p(\boldsymbol{\theta})$, using zero-centred Gaussian priors. Predictions therefore marginalise over possible parameter values:
\begin{equation}
    p(y\mid\mathbf{x},\mathcal{T})
    =
    \int
    p(y\mid\mathbf{x},\boldsymbol{\theta})
    p(\boldsymbol{\theta}\mid\mathcal{T})
    \,d\boldsymbol{\theta},
\end{equation}%
where $\mathcal{T}$ denotes the training data. For hierarchical models, $y$ corresponds to a valid path $\pi\in\Pi(\mathcal{H})$, whose probability is obtained from the conditional predictions along the hierarchy. Because the exact posterior $p(\boldsymbol{\theta}\mid\mathcal{T})$ is intractable for the neural networks considered here, the Bayesian models are trained using Stochastic Variational Inference (SVI)~\cite{hoffman2013stochastic}. An approximate posterior
$q_{\boldsymbol{\phi}}(\boldsymbol{\theta})$
is learned by maximising the Evidence Lower Bound (ELBO):
\begin{equation}
    \mathcal{L}_{\mathrm{ELBO}}
    =
    \mathbb{E}_{q_{\boldsymbol{\phi}}(\boldsymbol{\theta})}
    \left[
        \log
        p(\mathbf{y}\mid\mathbf{X},\boldsymbol{\theta})
    \right]
    -
    D_{\mathrm{KL}}
    \left[
        q_{\boldsymbol{\phi}}(\boldsymbol{\theta})
        \parallel
        p(\boldsymbol{\theta})
    \right].
\end{equation}%

The first term encourages the sampled networks to explain the observed behaviours, while the KL-divergence term regularises the learned parameter distribution toward the prior. For hierarchical Bayesian models, the same hierarchy-based penalty used in the deterministic formulation is added to the optimisation objective.

\subsection{Explanation Generation for Behaviour Selection}
\label{explanation_model}

The hierarchy exposes intermediate predictions, but the neural network itself does not directly provide feature-level explanations for these predictions. We therefore use decision-tree surrogate models~\cite{molnar2020interpretable} to approximate the learned decision functions. For the flat architecture, we fit a single surrogate tree $T_{\mathrm{flat}}$ using the conversational features as inputs and the flat neural network's predictions as targets:
\begin{equation}
    T_{\mathrm{flat}}:
    \mathbf{x}
    \rightarrow
    \hat{y}_{\mathrm{flat}},
\end{equation}%

For the hierarchical architecture, a separate surrogate tree is fitted to each prediction head. For a hierarchical decision node $v$, a surrogate tree can be denoted as 
$    T_v:
    \mathbf{x}
    \rightarrow
    \hat{y}_v.$ 
Thus, rather than approximating only the final joint prediction, the surrogate representation preserves the intermediate decisions exposed by the hierarchy. The corresponding surrogate decision paths can be written as follows:
\begin{equation}
    \rho_v(\mathbf{x})
    =
    \{
    r_{v,1},r_{v,2},\ldots,r_{v,m}
    \},
\end{equation}%
where each $r_{v,j}$ denotes a feature-based split encountered along the decision-tree path. For a hierarchical prediction, the paths associated with the relevant prediction heads are combined, providing a structured representation of the factors associated with the model's intermediate and final decisions:
\begin{equation}
    \mathcal{R}(\mathbf{x})
    =
    \bigcup_{k=0}^{K-1}
    \rho_{v_k}(\mathbf{x}).
\end{equation}


For the interpretability experiments, the extracted decision paths are provided to GPT-5.4, which converts them into natural-language explanations. In the flat condition, the LLM receives the single surrogate path associated with the final joint behaviour prediction, for example explaining why the robot selected a particular dialogue act--addressee combination. In the hierarchical condition, it instead receives all surrogate paths associated with the prediction heads traversed along the selected hierarchy, allowing the explanation to reflect the sequence of intermediate decisions leading to the final behaviour. For example, separate paths may describe why the robot selected an initiative act, then a question, and finally the whole group as addressee. The system prompt additionally provides the feature definitions and task context, including the definitions of the dialogue-act and addressee categories -- full prompt provided in~\ref{prompt_app}. The same prompt and explanation-generation procedure are used for both conditions, allowing us to investigate whether exposing the intermediate hierarchical decision structure leads to more interpretable explanations.
Examples of generated explanations along with the decision paths of the surrogate decision tree model(s) are available in Appendix \ref{app:explanation-examples}.
Algorithm~\ref{alg:hibridge} summarises hierarchical Bayesian behaviour prediction together with the subsequent post-hoc explanation-generation procedure used in the interpretability study.

\begin{algorithm}[t]
\caption{HiBRIDGE Hierarchical Bayesian Behaviour Prediction and Explanation Generation Pipeline.}
\label{alg:hibridge}
\begin{algorithmic}[1]
\Require Current conversation history $\mathcal{C}_t$

\vspace{1mm}
\Statex \hspace{3em} \textbf{--- Speak-or-Silence Prediction ---}
\State Encode $\mathcal{C}_t$ as conversational feature vector $\mathbf{x}_t$
\State Predict whether the robot should speak: 
\vspace{-0.5em}
\[
    \hat{s}_t
    \gets
    \arg\max_{s\in\{0,1\}}
    p(s\mid\mathbf{x}_t)
\]
\vspace{-1.8em}
\If{$\hat{s}_t = 0$}
    \State \Return \textsc{Remain Silent}
\EndIf

\vspace{2mm}
\Statex \hspace{2em} \textbf{--- Conversational Behaviour Prediction ---}

\State Propagate $\mathbf{x}_t$ through the hierarchical Bayesian network
\State Compute the posterior predictive probability of each valid path:
\vspace{-0.5em}
\[
p(\pi \mid \mathbf{x}_t,\mathcal{T})
\approx
\mathbb{E}_{q_{\boldsymbol{\phi}}(\boldsymbol{\theta})}
\left[
\prod_{k=1}^{K}
p_{\boldsymbol{\theta}}
(v_k \mid v_{k-1},\mathbf{x}_t)
\right]
\]%
\vspace{-0.5em}
\State Select the most probable behaviour path:
\vspace{-0.5em}
\[
\hat{\pi}_t
\gets
\arg\max_{\pi\in\Pi(\mathcal{H})}
p(\pi\mid\mathbf{x}_t,\mathcal{T})
\]%
\vspace{-1em}
\State Decode $\hat{\pi}_t$ into dialogue act $\hat{d}_t$ and addressee $\hat{a}_t$

\vspace{4mm}
\Statex \hspace{3.5em} \textbf{--- Explanation Generation ---}

\State Extract and combine the surrogate decision paths:
\[
\mathcal{R}(\mathbf{x}_t)
\gets
\bigcup_{k=0}^{K-1}
\rho_{v_k}(\mathbf{x}_t)
\]
\State Generate explanation: $e_t \gets \mathrm{LLM}\!\left(\mathcal{R}(\mathbf{x}_t)\right)$

\State \Return $(\hat{d}_t,\hat{a}_t,e_t)$
\end{algorithmic}
\end{algorithm}

\section{Offline Dataset Experiments and Results}

This section presents the offline group-HRI datasets, baseline methods, and results across these datasets, comparing hierarchical and flat variants of HIBRIDGE against several state-of-the-art baselines, thereby supporting our \textbf{Contribution~1}.

\subsection{Offline Datasets}
In our work, we selected datasets based on three inclusion criteria: the interactions had to involve a conversation, include at least one robot, and involve at least two people. Within this scope, we chose three datasets representing distinct interaction contexts in which robot behaviour differs substantially: (i) \textbf{MuMMER dataset}~\cite{canevet2020mummer} where a robot entertains groups of people in a shopping mall, (ii) \textbf{Addlesee dataset}~\cite{addlesee2023data} where a robot provides information in hospitals, and (iii) \textbf{Spitale dataset} where a robot moderates a group negotiation task~\cite{spitale2026understanding}.

First, the \emph{MuMMER dataset}~\cite{canevet2020mummer} consists of 33 conversations (1,733 speaking turns) involving two or three participants interacting with a robot in a shopping mall environment. The robot engages users through jokes, quizzes, news updates, and directions. The dataset includes video, audio, and sensor data; however, transcripts were generated from video using the Whisper model (turbo variant)~\cite{radford2023robust} and subsequently corrected manually. Second, the \emph{Addlesee dataset}~\cite{addlesee2023data} contains 30 conversations (1,050 speaking turns) between two participants and a robot in a memory clinic setting. The robot responds to patient questions about the facility and appointments. This dataset is text-only, with addressee annotations already provided. 
Third, the \emph{Spitale dataset} \cite{spitale2026understanding} comprises 28 conversations (1,043 speaking turns) in which two or three participants interact with two robots, which are treated as a single conversational agent for the purposes of this work. The robots facilitate a negotiation game in which participants decide how and at what price to sublet rooms in a shared apartment. This dataset is also text-only, with addressee annotations available for robot utterances only.
When annotations were not present, a human annotator (the first author of the paper)  labelled dialogue acts and addressees for all utterances across datasets, covering both robot and participant turns, to support feature extraction.
Addressee was determined by whom the utterance's content was directed toward, e.g. a name, a direct question, or content clearly relevant to a single participant, and labelled as \texttt{ALL} whenever the utterance was phrased to include the whole group rather than one individual. Dialogue acts were assigned by matching each utterance against the category definitions given in our taxonomy (Section~\ref{sec:dialogue_act_taxonomy}).

Although the MuMMER dataset originally included video, it was collected for person re-identification, and the camera motion made visual features unstable. Preliminary experiments using audio features did not yield improvements over text-based conversational structure features. For this reason, we restrict all analyses to textual data across datasets.

\subsection{Baselines for Offline Datasets}
The baselines we chose are two LLM baselines, deepseek-chat \footnote{\url{https://www.deepseek.com/}} and gpt-4.1-mini \footnote{\url{https://platform.openai.com/docs/models}}, as well as a supervised learning approach, i.e., a state-of-the-art model in multi-party human–robot interaction (Gillet et al \cite{gillet2025templates}). 

\textbf{LLMs}: For the LLM baselines, we adopted a prompt structure inspired by Addlesee et al. \cite{addlesee2024multi}.
Each prompt included: (i) an explanation of the dataset context and the prediction task (addressee and dialogue act classification); (ii) a specification of the valid labels for addressees and dialogue acts, along with their definitions; (iii) the dialogue history; (iv) a list of previously selected answers to avoid; and (v) three examples. The models were asked to predict a joint combination of addressee and dialogue act for each prompt.
The dialogue history was truncated to the five most recent turns. All LLMs were queried with a temperature of 1, top-p of 1, and a maximum output length of 40 tokens.

\textbf{Gillet et al.}: For the Gillet et al. \cite{gillet2025templates} baseline, we used the architecture provided in the authors' public GitHub repository \footnote{\url{https://github.com/sarahgillet/TGM-SmallGroups}} and adapted our dataset and feature representations to match their framework. This required design choices to align our computed features and annotations with those used in their work.
Gillet et al.’s approach \cite{gillet2025templates}, based on a Graph Neural Network (GNN), organises features into three categories: node features, representing individual participants; edge features, capturing interactions between participants; and global features, encoding group-level information. For node features, we included the total time spoken by each participant and the total time they were addressed. For edge features, we included the amount of time one participant addressed another. All remaining features, including speaker identity, addressee, and dialogue act labels from previous turns, and robot speaking and addressing times, were treated as global features.
In Gillet et al., the task of deciding who to address and what to say is divided between two models. For addressee prediction, a GNN outputs a binary decision (addressed vs. not addressed) for each participant, assuming only individual addressees rather than a group-level label. To accommodate this, during training, we converted instances annotated with addressing the whole group by duplicating them once for each participant. At inference time, we followed the rule described in Gillet et al. \cite{gillet2025templates}: if all participants were predicted as being addressed, the addressee was set as the whole group. 
To make the Gillet et al.~\cite{gillet2025templates} baseline comparable to our models, we retrained their entire architecture from scratch, using the leaf-level classes of our own dialogue-act taxonomy (Section~\ref{sec:dialogue_act_taxonomy}) as the classes for the what-to-say component, instead of the original ten classes describing the robot's social behavior toward the addressed participant $p_{who}$ (e.g., agree with $p_{who}$, ask $p_{who}$ to elaborate, praise $p_{who}$).
\subsection{Implementation Details}
\label{imp_details}
\textbf{Data Augmentation:} To reduce overfitting to the limited interaction data, we augment the training representations by perturbing randomly selected numerical features with Gaussian noise~\cite{machado2022benchmarking}. For a numerical feature $x_j$, an augmented value is given by
\[
    x'_j = x_j + \epsilon_j,
    \qquad
    \epsilon_j \sim \mathcal{N}(0,\sigma_{\mathrm{aug}}^2),
\]%
where $\sigma_{\mathrm{aug}}$ controls the magnitude of the perturbation and is selected during hyperparameter optimisation.

\textbf{Training Procedure:}  
All datasets were partitioned into ten randomly stratified folds to preserve the distribution of dialogue act–addressee combinations across splits. Within each fold, the training portion was further divided into 80\% training and 20\% validation sets, maintaining stratification. This results in an overall split of 72\% training, 18\% validation, and 10\% test data.

Deterministic models were implemented in PyTorch and trained using Adam for up to 1000 epochs, with early stopping using a patience of 15 epochs. The Bayesian architectures were implemented in Pyro~\cite{bingham2019pyro} and trained for up to 1000 epochs, using early stopping with a patience of 100 epochs. Gradient clipping and learning-rate decay were used to stabilise optimisation. For the Gillet et al. baseline, as in our other experiments, models were trained for up to 1000 epochs with early stopping using a patience of 15 epochs. 

\textbf{Hyperparameter Optimisation:} For all the models used in this work, we performed hyperparameter tuning using Optuna~\cite{akiba2019optuna}, with 75 trials per model. We selected hyperparameters based on average validation accuracy obtained via 10-fold cross-validation on the largest dataset used in our experiments, the MuMMER dataset. The hyperparameter search spaces varied by type of model. 

For our proposed \emph{deterministic models}, the tuned hyperparameters included the number of hidden layers, the number of neurons per layer, dropout rate, L2 regularisation strength, the choice and weighting of the loss function (including class-weighted cross-entropy), the number of augmented data copies, the magnitude of injected noise, and the penalty coefficient used in the hierarchical loss.

For our proposed \emph{Bayesian models}, the tuned hyperparameters included the number of hidden layers and neurons per layer, dropout rate, learning rate, the standard deviations of the priors for weights and biases, the number of augmented data copies, the magnitude of injected noise, the weighting of the cross-entropy component, and the penalty coefficient used in the hierarchical loss.

For our proposed model that decides whether the robot should \emph{speak or remain silent}, the hyperparameter search space included the number of hidden layers, the number of neurons per layer, dropout rate, L2 regularisation strength, the number of augmented data copies, and the magnitude of the injected noise.

For the \emph{Gillet et al. baseline} \cite{gillet2025templates} we followed the parameter ranges specified by the hyperparameter tuning the original paper, with the addition of data augmentation parameters used previously, namely the number of dataset replications and the amount of noise injected into feature columns.

Full hyperparameter configurations and model weights are available in the~\ref{param_app}.

\subsection{Evaluation Procedure on Offline Datasets}
\label{mettics}
The offline evaluation is designed to answer three central questions. First, we assess whether the proposed models can outperform the baseline approaches, establishing that if any observed gains are statistically significant. Second, we investigate the impact of formulating the neural networks within a Bayesian framework, to determine whether this yields a measurable improvement in overall performance. Finally, we examine whether hierarchical architectures provide any advantage over a flat formulation, and which hierarchical ordering leads to better performance.
To investigate these questions, we evaluated three architectural paradigms: flat classification, hierarchical addressee-first, and hierarchical dialogue-act-first, each implemented with both Bayesian neural networks and standard (non-Bayesian) neural networks, resulting in a total of six model variants.
%

\textbf{Metrics:} Model performance was evaluated across all three datasets on the test splits obtained from 10-fold stratified cross-validation. We report the accuracy in Hit@1, Hit@3, and Hit@5, measuring whether the correct behaviour appears among the model’s top 1, 3, or 5 predictions, respectively. These metrics are standard for similar tasks such as utterance selection \cite{park2022bert} and are well-suited to group conversations, where multiple combinations of dialogue acts and addressees may be equally valid.

\textbf{Statistical Testing:} To assess whether the observed differences among proposed and baseline models are statistically significant, we conduct pairwise statistical comparisons between every proposed model and each baseline. Specifically, we apply Welch's t-tests~\cite{welch1947generalization} when normality assumptions are satisfied, and Mann–Whitney U tests~\cite{mann1947test} otherwise, evaluating performance on Hit@1, Hit@3, and Hit@5 (resulting in three comparisons per proposed model). All p-values are adjusted using the Benjamini–Hochberg correction~\cite{ferreira2006benjamini}.

Further, to isolate the individual contributions of the Bayesian and hierarchical design choices, and specifically to determine whether the Bayesian formulation yields a significant positive effect on performance, we perform a 2×3 factorial ANOVA. The two factors are model type (deterministic vs. Bayesian) and hierarchical structure (flat, hierarchical with addressee-first, and hierarchical with dialogue-act-first); this factorial design lets us disentangle the effect of the Bayesian formulation from that of the hierarchical ordering, rather than conflating the two.
This analysis is conducted across all datasets and evaluation metrics. When the assumptions of normality and homogeneity of variance are violated, we instead use the Scheirer–Ray–Hare test~\cite{scheirer1976analysis}. 
A 2×3 factorial ANOVA serves as the primary framework, after which the effects of the Bayesian and hierarchical factors are examined separately only if they are significant in the 2×3 factorial ANOVA. To examine the individual effect of a significant factor, we perform post hoc pairwise comparisons using Welch's $t$-test, or the Mann–Whitney $U$ test when normality is violated. When the Bayesian factor is significant, these comparisons are run between the deterministic and Bayesian variants within each architecture, for every evaluation metric. When the hierarchical architecture factor is significant, comparisons are instead run pairwise between the three architectures (flat, hierarchical addressee-first, and hierarchical dialogue-act-first) within each modelling type (deterministic and Bayesian), with correction for multiple comparisons applied given the three pairwise tests involved.

\subsection{Results on  Offline Datasets}

\begin{table*}[t]
\centering
\caption{Offline evaluation results across Mummer \cite{canevet2020mummer}, Addlesee \cite{addlesee2023data}, and Spitale \cite{spitale2026understanding} datasets.
Top row shows mean Hit@k, bottom row shows standard deviation in parentheses; bold marks the best result per column.
$^{*}p<.05$, $^{**}p<.01$, $^{***}p<.001$ (Welch's $t$-test, or Mann--Whitney $U$ where normality/variance assumptions were violated; Benjamini-Hochberg corrected) indicate that a proposed model significantly outperforms all three baselines, reflecting the least significant of the three comparisons}
\label{tab:offline_results_all}
\renewcommand{\arraystretch}{1.15}
\setlength{\tabcolsep}{3pt}
\scriptsize

\begin{tabular}{|c|l|ccc|ccc|ccc|}
\hline
& \multirow{2}{*}{\textbf{Model}} 
& \multicolumn{3}{c|}{\textbf{Mummer} \cite{canevet2020mummer}} 
& \multicolumn{3}{c|}{\textbf{Addlesee} \cite{addlesee2023data}} 
& \multicolumn{3}{c|}{\textbf{Spitale} \cite{spitale2026understanding}} \\
\cline{3-11}
&
& Hit@1 & Hit@3 & Hit@5
& Hit@1 & Hit@3 & Hit@5
& Hit@1 & Hit@3 & Hit@5 \\
\hline

\multirow{6}{*}{\rotatebox[origin=c]{90}{\textbf{Baselines}}}
& deepseek-chat~\cite{addlesee2024multi}
& 0.2051 & 0.2605 & 0.3194
& 0.3056 & 0.3513 & 0.4014
& 0.1208 & 0.2226 & 0.2758 \\
&
& (0.0116) & (0.0124) & (0.0161)
& (0.0205) & (0.0295) & (0.0238)
& (0.0232) & (0.0285) & (0.0230) \\
\cline{2-11}

& gpt-4.1-mini~\cite{addlesee2024multi}
& 0.1525 & 0.2946 & 0.3571
& 0.1880 & 0.2489 & 0.2816
& 0.0723 & 0.1849 & 0.2628 \\
&
& (0.0049) & (0.0064) & (0.0110)
& (0.0200) & (0.0198) & (0.0213)
& (0.0186) & (0.0220) & (0.0269) \\
\cline{2-11}

& Gillet et al.~\cite{gillet2025templates}
& 0.4781 & 0.5708 & 0.6464
& 0.5781 & 0.6250 & 0.7312
& 0.4906 & 0.5625 & 0.5844 \\
&
& (0.0312) & (0.0497) & (0.0509)
& (0.0646) & (0.0489) & (0.1044)
& (0.1063) & (0.1179) & (0.1122) \\
\hline

\multirow{12}{*}{\rotatebox[origin=c]{90}{\textbf{Ours}}}
& Flat NN
& 0.5494$^{***}$ & 0.6969$^{***}$ & 0.7683$^{***}$
& 0.7408$^{***}$ & 0.8621$^{***}$ & 0.9175$^{***}$
& 0.5987$^{*}$ & 0.7494$^{***}$ & 0.8246$^{***}$ \\
&
& (0.0319) & (0.0324) & (0.0254)
& (0.0355) & (0.0464) & (0.0272)
& (0.0659) & (0.0634) & (0.0616) \\
\cline{2-11}

& Hierarchical NN (DACT$\rightarrow$ADDR)
& 0.532$^{***}$ & 0.6975$^{***}$ & 0.7613$^{***}$
& 0.7621$^{***}$ & 0.8650$^{***}$ & 0.8835$^{***}$
& 0.555 & 0.6819$^{*}$ & 0.7751$^{***}$ \\
&
& (0.0261) & (0.0290) & (0.0192)
& (0.0324) & (0.0362) & (0.0355)
& (0.0561) & (0.0746) & (0.0648) \\
\cline{2-11}

& Hierarchical NN (ADDR$\rightarrow$DACT)
& 0.5756$^{***}$ & 0.7225$^{***}$ & 0.7858$^{***}$
& 0.7583$^{***}$ & 0.8903$^{***}$ & 0.9039$^{***}$
& 0.5937$^{*}$ & 0.7146$^{**}$ & 0.8068$^{***}$ \\
&
& (0.0352) & (0.0241) & (0.0263)
& (0.0435) & (0.0311) & (0.0244)
& (0.0582) & (0.0570) & (0.0727) \\
\cline{2-11}

& Flat Bayesian NN
& \textbf{0.6396}$^{***}$ & \textbf{0.8137}$^{***}$ & \textbf{0.8409}$^{***}$
& \textbf{0.8019}$^{***}$ & \textbf{0.9087}$^{***}$ & \textbf{0.9184}$^{***}$
& \textbf{0.678}$^{***}$ & 0.8158$^{***}$ & 0.8415$^{***}$ \\
&
& (0.0616) & (0.0472) & (0.0282)
& (0.0530) & (0.0334) & (0.0279)
& (0.0936) & (0.1131) & (0.0934) \\
\cline{2-11}

& Hierarchical Bayesian NN (DACT$\rightarrow$ADDR)
& 0.596$^{***}$ & 0.7748$^{***}$ & 0.8195$^{***}$
& 0.7864$^{***}$ & 0.8922$^{***}$ & 0.9097$^{***}$
& 0.6502$^{**}$ & 0.8148$^{***}$ & 0.8504$^{***}$ \\
&
& (0.0488) & (0.0517) & (0.0388)
& (0.0434) & (0.0395) & (0.0330)
& (0.0804) & (0.1048) & (0.0744) \\
\cline{2-11}

& Hierarchical Bayesian NN (ADDR$\rightarrow$DACT)
& 0.632$^{***}$ & 0.8073$^{***}$ & 0.8294$^{***}$
& 0.7816$^{***}$ & 0.9019$^{***}$ & 0.9126$^{***}$
& 0.671$^{**}$ & \textbf{0.8336}$^{***}$ & \textbf{0.8554}$^{***}$ \\
&
& (0.0682) & (0.0611) & (0.0441)
& (0.0437) & (0.0319) & (0.0263)
& (0.0799) & (0.0847) & (0.0732) \\
\hline

\end{tabular}
\end{table*}

\subsubsection{Proposed Models versus Baselines}

As shown in Table~\ref{tab:offline_results_all}, the proposed models generally achieve higher mean performance than the baseline methods across all evaluation metrics and datasets. Pairwise Welch's $t$-tests (or Mann--Whitney $U$ tests when normality assumptions were violated) confirm that the proposed approaches consistently outperform the baseline methods by a statistically significant margin (all $p < 0.05$). The only exception occurs for the hierarchical dialogue-act-first model (\textit{Hierarchical NN (DACT$\rightarrow$ADDR)}) on the Hit@1 metric in the Spitale dataset when compared to the Gillet et al. baseline; however, the proposed model still achieves a higher mean performance.
This indicates that, overall, our proposed models are well-suited for this task.

From the performance of our best-performing models, we observe that Hit@1 scores reach 0.6396 for MuMMER, 0.678 for Spitale and 0.8019 for Addlesee. 
Given that the Addlesee dataset achieves the highest overall performance, the task of learning appropriate robot behaviour in this setting can be considered comparatively easier than in the other datasets. This interpretation is further supported by the nature of the interaction scenarios: in the Addlesee setting, the robot assumes a relatively passive role, primarily responding to participants' questions, whereas the Spitale and MuMMER datasets involve more dynamic and demanding behaviours---in Spitale, the robot (or pair of robots) actively manages the flow of the conversation, while in MuMMER the robot must handle a wider range of functions, including entertaining users with quizzes, maintaining small talk, and answering users' questions.

\subsubsection{Effects of Bayesian and Hierarchical Modelling}

To examine the respective contributions of Bayesian modelling and hierarchical architecture to predictive performance, we analysed the effects of the two design choices using the 2$\times$3 factorial ANOVA or Scheirer--Ray--Hare test described in Section~\ref{mettics}. In general, these tests do not show any significant effect of hierarchical modelling for any evaluation metric. In contrast, they frequently indicate that Bayesian modelling has a significant positive effect on predictive performance, although the strength of this effect differs depending on the dataset and metric.

For the Spitale and MuMMER datasets, the pattern is consistent: Bayesian modelling always has a significant effect on predictive performance (MuMMER: Hit@1 $H = 19.75$; Hit@3 $F = 29.38$; Hit@5 $F = 51.50$; Spitale: Hit@1 $H = 18.48$; Hit@3 $F = 23.12$; Hit@5 $F = 6.02$; all $p < .05$). To further examine this effect, we conducted post hoc comparisons (Welch's $t$-test, or the Mann–Whitney $U$ test when normality was violated) between the deterministic and Bayesian variants of each architecture, for every metric. The only exception to this trend occurred at the Hit@5 metric in the Spitale dataset, where Bayesian neural networks were not significantly better than their deterministic counterparts for the flat ($U = 61.5$, $p = .202$) and hierarchical addressee-first ($U = 72.0$, $p = .052$) architectures. This suggests that as the task becomes easier (i.e., when multiple attempts are allowed), the advantage of Bayesian modelling may diminish, even though it remains important for more challenging settings such as Hit@1 and Hit@3.

A similar but more pronounced trend is observed in the Addlesee dataset. The 2×3 ANOVA (or the Scheirer–Ray–Hare test when assumptions are violated) indicates a significant effect of Bayesian modelling for Hit@1 ($F = 10.94$, $p = .002$) and Hit@3 ($F = 8.99$, $p = .004$), but not for Hit@5 ($F = 2.50$, $p = .119$). Post hoc comparisons further show that differences between Bayesian and deterministic models are significant only for the flat architecture at Hit@1 and Hit@3.

Together, these results suggest that the advantage of Bayesian modelling may diminish as task difficulty decreases. This pattern is consistent with the comparatively higher performance observed in the Addlesee dataset and with the reduced Bayesian advantage at Hit@5, where multiple predictions are accepted.

\section{Online Interpretability Study and Results}
This section presents an interpretability study designed to assess whether our hierarchical architecture yields tangible benefits for explainability of our system, supporting our \textbf{Contribution 2}. This section details the study design and presents its results.

\subsection{Online Study Design}


\subsubsection{Hypotheses}




Prior work suggests that decomposing a model into modular components and exposing intermediate outputs can improve interpretability~\cite{swamy2023multimodn, hu2018explainable}, while hierarchically structured explanations have been perceived as more interpretable than flat ones~\cite{chen2020generating}. We therefore expect explanations
from the hierarchical model to be more interpretable than those from the flat model. Specifically, we hypothesise that hierarchical explanations will be perceived as more understandable, coherent, and useful (\textbf{H1}), and accordingly participants will prefer hierarchical over flat explanations when directly comparing the two (\textbf{H2}).

Beyond perceived interpretability, prior work shows that explanations of robot behaviour can help people predict subsequent robot behaviour~\cite{love2024would}. Together with our expectation that hierarchical explanations are more interpretable than flat explanations (H1), we hypothesise that participants will achieve higher accuracy in inferring robot behaviour when given hierarchical rather than flat explanations (\textbf{H3}). More generally, we hypothesise that participants will achieve higher accuracy when aided by an explanation than when no explanation is provided (\textbf{H4}).

\subsection{Online Study Participants and Procedure}
Having established in \textbf{Contribution 1} that our proposed models outperform state-of-the-art baselines, we turn in the online evaluation to assessing their interpretability. In particular, we investigated whether explanations derived from a hierarchical model provide greater clarity and usefulness than those from a flat model. To this end, we conducted a within-subjects online study to evaluate the quality of explanations corresponding to robot behaviour in recorded interaction clips from the in-person study.

The study was conducted using an online questionnaire hosted on Qualtrics, with participants recruited via Prolific.
In an initial pilot phase, six individuals recruited via personal contacts completed an early version of the questionnaire. Their feedback was used to refine and improve the study design and survey structure.
Following finalisation of the questionnaire, a sample of 20 participants was recruited through Prolific. Eligibility criteria included English as a first language, age over 18, and a minimum approval rating of 99\%. We also selected a balanced distribution of male and female participants (i.e., 10M:10F). The questionnaire took approximately 30 minutes to complete, and each participant was compensated according to an hourly rate of £12. 
The study consisted of three main tasks. 

\subsection{Online Study Tasks}
\label{online_taks}
To evaluate our hypotheses, we structured the online study into three tasks. In each task, participants watched video clips depicting a robot interacting with two people (the videos were obtained from the in-person study described in the next section). The interactions were inspired by the MuMMER dataset, in which participants were instructed to imagine themselves in a shopping mall while the robot interacted with them by providing directions, asking questions for entertainment, and engaging in small talk.
Participants watched a total of 14 clips, each ranging from 30 to 50 seconds in length. Each clip included all the contextual information available to the model when making its decisions, consisting of a maximum of the five preceding turns of conversation.


\textbf{Task 1} aimed to obtain a qualitative assessment of explanation quality. Participants viewed, in a counterbalanced order, two blocks of three clips (one per model: flat or hierarchical). In each clip, participants were shown an explanation of the robot’s behaviour, and after watching it, they completed six custom-designed questions evaluating explanation quality (shown in~\ref{ques_app}). These questions assessed understandability, perceived usefulness for interpreting the robot’s behaviour, the extent to which the explanation clarified the selected dialogue act and addressee, and the coherence between the explanation and the robot’s choices of dialogue act and addressee within the context of the clip. At the end of each block, participants completed the validated Explanation Satisfaction Scale (ESS) questionnaire \cite{hoffman2023measures}.

\textbf{Task 2} was designed to provide a more objective measure of understanding and to assess whether explanations supported participants’ ability to infer the robot’s behaviour. Participants watched four clips depicting interactions that paused immediately before the robot’s intervention. After each clip, they were shown an explanation of the robot’s subsequent decision, while the specific addressee (individual or group) and intended dialogue act were masked. Participants were then asked to infer the missing information by indicating whom the robot would address and what it would say (selected from a predefined set of dialogue acts). Of the four clips, two were paired with explanations generated by the hierarchical model and two with explanations generated by the flat model. Both the assignment of model type to clips and the order of presentation were randomised and counterbalanced across participants.

\textbf{Task 3} evaluated whether participants had learned the robot’s behavioural patterns and enabled a direct comparison between hierarchical and flat explanations for the same interaction context. As in the second task, participants viewed four clips in which interactions paused immediately before the robot’s intervention. However, in this case, they were first asked to predict the robot’s decision without any explanatory information. They were then shown two full explanations for the same interaction—one generated by the hierarchical model and one by the flat model—and asked to indicate which explanation better aligned with the robot’s behaviour. The order of explanation presentation was randomised and counterbalanced across participants. To control for potential order effects, the sequence of the second and third tasks was also randomised across participants.
Examples of the tasks in the online study can be found in the appendix \ref{app:qualtrics_screenshots}.

\subsection{Online Study Metrics}
To test \emph{H1}, we evaluated the results from Task 1. Specifically, we analysed the ESS scores by summing the responses for each item to obtain a total score, from which a mean score is calculated. We then compared the hierarchical and flat conditions using a Wilcoxon signed-rank test. Additionally, we apply Wilcoxon signed-rank tests to each item of the self-designed questions separately to examine which specific dimensions differ between conditions. To account for the multiple comparisons across these tests, we applied a Benjamini-Hochberg false discovery rate correction. These dimensions include explanation understandability, usefulness for interpreting the robot’s behaviour, the extent to which the explanation clarifies the selected dialogue act and addressee, and the coherence between the explanation and the robot’s choice of dialogue act and addressee within the context of the clip.

To test \emph{H2}, we analysed the results from Task 3. Specifically, we compared the number of times hierarchical or flat explanations are preferred across all questions for each participant. A paired t-test was conducted if the data met the normality assumption; otherwise, a Wilcoxon signed-rank test was applied.

To test \emph{H3}, we analysed the scores obtained from Task 2. Specifically, we compared participants’ accuracy in correctly selecting robot behaviour after viewing hierarchical versus flat explanations, using a paired t-test when the data was normally distributed; otherwise, we used a Wilcoxon signed-rank test.

To test \emph{H4}, we compared participants’ accuracy scores with and without explanations between Task 2 and Task 3. A paired t-test is used if the data are normally distributed; otherwise, a Wilcoxon signed-rank test is conducted.

\subsection{Online Study Results}


To evaluate \textbf{H1}, we compared participants' ratings of the Explanation Satisfaction Scale (ESS) as well as the scores from the self-designed questionnaire between the hierarchical and flat explanation conditions. 
Contrary to our expectations, although the hierarchical explanations received a higher average rating (\(M = 3.59\)) than the flat explanations (\(M = 3.34\)) no statistically significant difference was found in the ESS scores between the two conditions (\(p > .05\)).
One possible explanation is that the effect size is relatively small and that a larger sample would be required to detect a significant difference. 
A similar pattern emerged for the self-designed questionnaire. Overall, hierarchical explanations received slightly higher ratings (\(M = 4.28\)) than flat explanations (\(M = 4.18\)), but the overall difference was not statistically significant (\(p > .05\)).

\begin{figure}[t!]
    \centering
    \includegraphics[
        width=0.5\textwidth,
    ]{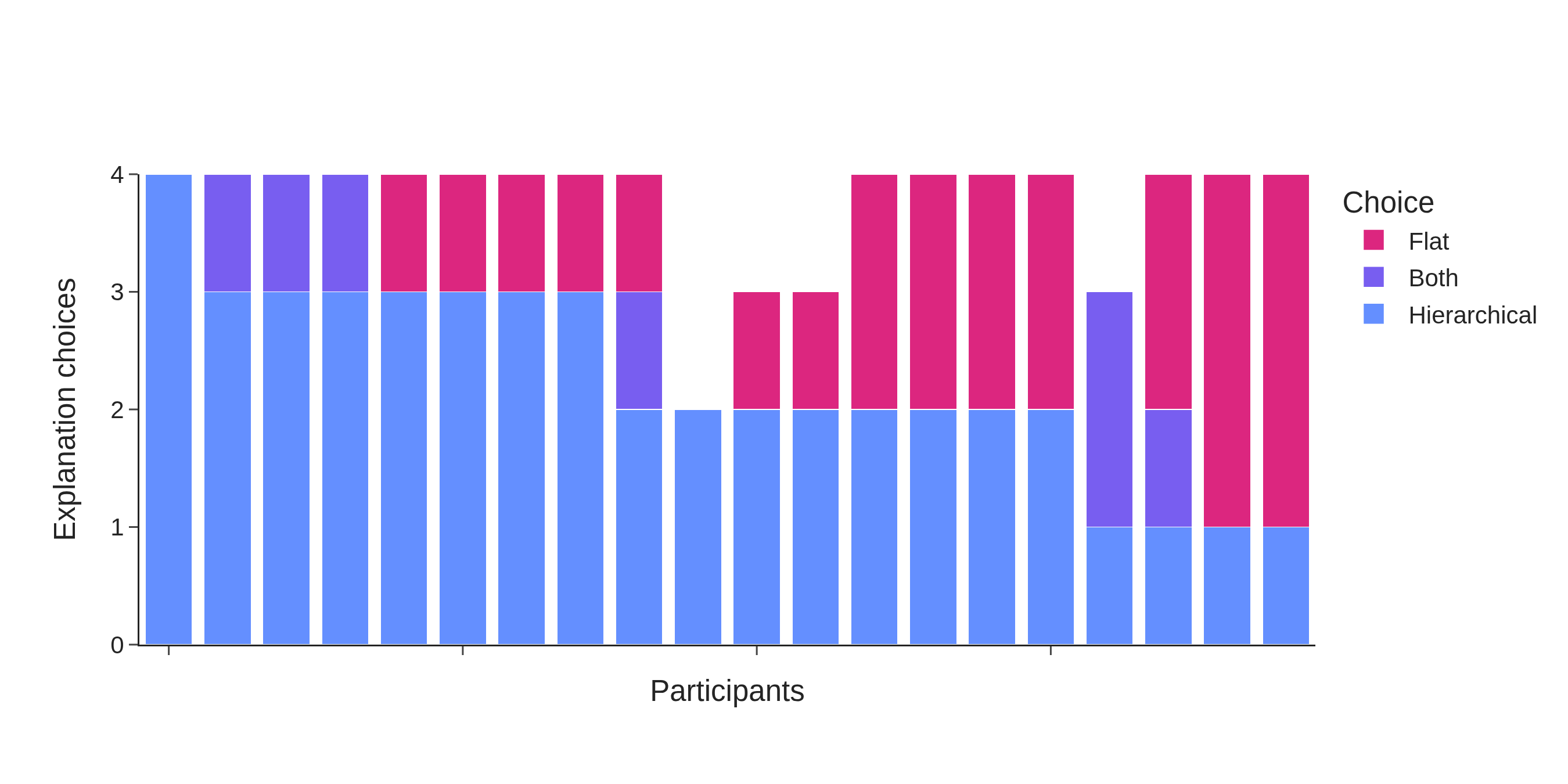}
    \caption{Participants' explanation preferences in Task 3 of the online interpretability study, where the hierarchical and flat explanations were shown side by side for each of four video clips. Each bar represents one participant; the stacked segments show how many of their four choices went to the hierarchical explanation, the flat explanation, or ``Both'' when they judged the two explanations equally good.}
    \label{online study}
\end{figure}


To evaluate \textbf{H2}, we examined participants' preferences when the hierarchical and flat explanations were presented side by side for the same clip; for each clip, participants could choose the hierarchical explanation, the flat explanation, both (if they judged the two equally acceptable), or neither. As we can see in Figure \ref{online study}, across 20 participants and four comparison clips each (80 judgements in total) the hierarchical explanation was chosen on 45 occasions and the flat explanation on 23, with both explanations judged equally acceptable in 7 cases and neither acceptable in 5. To test whether this difference was reliable, we computed for each participant how many of their four judgements favoured the hierarchical explanation versus the flat explanation, counting a ``both'' judgement toward each side; this gave per-participant means of M = 2.60 for hierarchical and M = 1.50 for flat. A Wilcoxon signed-rank test on these paired per-participant scores confirmed that hierarchical explanations were selected significantly more often than flat explanations (W = 21.00, p = .014). Together, these results support \textbf{H2}, indicating that participants preferred the hierarchical explanations over the flat ones.

To evaluate \textbf{H3} and \textbf{H4}, we analysed participants' accuracy in predicting the robot's decisions. No significant differences were found between the no-explanation, hierarchical-explanation, and flat-explanation conditions. Participants' prediction accuracies were comparable across all three conditions, indicating that neither hierarchical nor flat explanations improved participants' ability to infer the robot's future behaviour. One possible explanation is that the prediction task was inherently challenging, with several possible behaviours being plausible at the same time.

\section{In-person Study and Results}
This section describes the integration of our models into a real-time robotic system and its evaluation through an in-person user study aimed at assessing whether the system was well perceived by users, supporting our \textbf{Contribution 3}. We first detail the system implementation, then present the study protocol, and finally present its findings.

\subsection{Robotics Deployment}
\begin{figure*}[t]
    \centering
    \includegraphics[width=\textwidth]{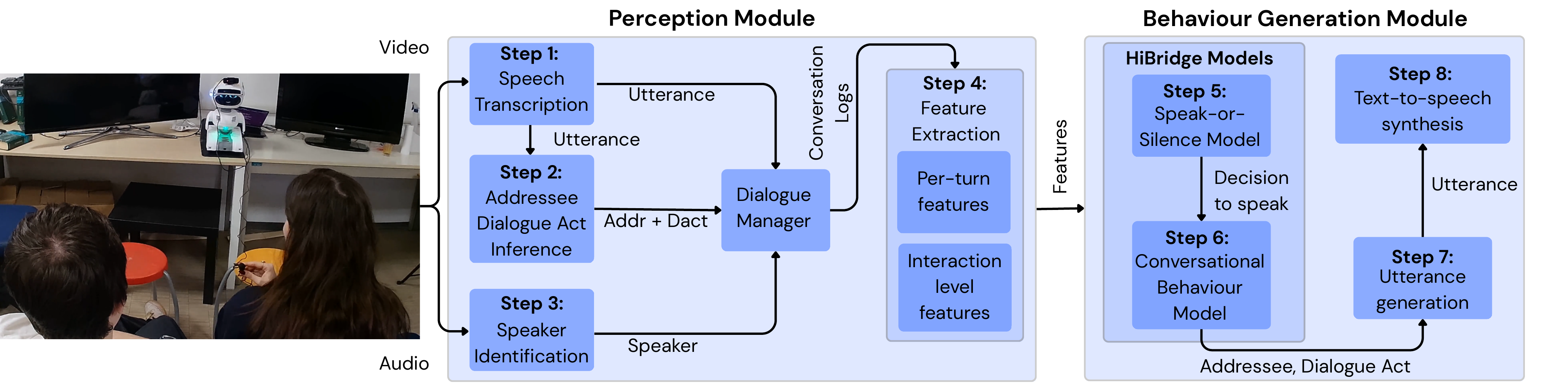}
    \captionsetup{justification=centering}

    \caption{System architecture used in the in-person study, following the HARMONI framework~\cite{spitale2021composing}. The Perception Module (right) transcribes speech, identifies speakers, infers each utterance's addressee and dialogue act, and extracts the features required by the HiBridge models. The Behaviour Generation Module (left) uses the HiBridge models to decide whether to intervene in the conversation and, if so, select the robot's addressee and dialogue act, then prompts an LLM to generate the resulting utterance, which is synthesised into speech.}
    \label{in person}
\end{figure*}

To integrate our trained models into a real-time system, we used the open-source HARMONI framework \cite{spitale2021composing}. The system was divided into two parts: a \textbf{perception module}, responsible for understanding the group conversation, followed by a \textbf{behaviour generation module}, responsible for producing the robot's responses. Alternation between the two modules was governed by fixed silence windows: a two-second silence indicated that a participant's turn had ended, while a three-second window following the robot's turn allowed another speaker to take the floor.
The \emph{perception module} performed the following steps: 
\begin{itemize}
    \item \textbf{Step 1 (Speech transcription):} Each utterance was transcribed in real time using the Google Speech-to-Text (STT) service, which also performed speaker diarization. A fixed one-second silence threshold was used to segment and finalise each utterance. 
    \item \textbf{Step 2 (Addressee and dialogue act inference):} For each participant utterance, GPT-4.1-mini was used to infer the addressee and dialogue act of the previous turn, conditioned on the conversation transcript so far.
    \item \textbf{Step 3 (Speaker identification):} Speaker identity was determined by combining the diarization output with facial-feature-based speaker identification using dlib \cite{king2009dlib}.
    \item \textbf{Step 4 (Feature extraction):} The features required by our models were computed from the current state of the conversation transcript.
\end{itemize}

The \emph{behaviour generation module} then performed the following steps: 
\begin{itemize}
\item \textbf{Step 5 (Speak-or-Silence Model):} A binary classifier trained offline determined whether the robot should speak at all (\emph{HiBRIDGE Speak-or-Silence Model}). 
\item \textbf{Step 6 (Conversational Behaviour  Model):} If the classifier decided the robot should speak, the previously trained offline models were used to select the robot's addressee and dialogue act (\emph{HiBRIDGE Conversational Behaviour Model}). 
\item \textbf{Step 7 (Utterance generation):} A large language model (specifically GPT-4.1-mini) was then prompted with the predicted dialogue act, the selected addressee, and the list of current speakers to generate the robot's utterance. If the addressee was not the entire group, GPT-4.1-mini was also asked to explicitly resolve the intended individual addressee from the list of available speakers. 
\item \textbf{Step 8 (Text-to-speech synthesis):} The generated utterance was then converted into speech using the voice "Amy" from Amazon Polly, which the robot used to deliver the response.
\end{itemize}

Figure~\ref{in person} shows the overall architecture used in the in-person study. The architecture described above is platform-independent; we now describe the specific platform used in our in-person study. The robot platform used was a Misty II\footnote{\texttt{https://www.mistyrobotics.com/misty-ii}}, which provided three actuators to support the interaction: head movement, arm movement, and a light on its chest. 
The robot's gaze depended on the selected addressee: it oriented its head towards the addressed participant's face when speaking to an individual, and towards the midpoint between participants when addressing the group. 
Arm movement was included to add a degree of physical movement to the interaction. We handcrafted a small set of fixed arm gestures, each mapped to a specific dialogue act (e.g.\ raising both arms at different angles, raising a single arm); these gestures were not designed to be meaningfully distinguishable from one another and served purely to introduce movement rather than to convey additional information. Finally, the robot's chest light indicated its internal state to participants, who were informed of this convention beforehand: it was green while the perception module was active (i.e.\ while the robot was listening) and red while the behaviour generation module was active (i.e.\ while the robot was speaking).
\subsection{In-Person Study Participants and Procedure}
In the in-person study, we examined whether our models, in combination with an LLM, could adequately manage the robot’s behaviour in a group conversation setting. The study was approved by the institutional ethics committee of Politecnico di Milano. We recruited a total of 12 participants (7 male, 5 female), aged 24–35, from a convenience sample of PhD students, postdoctoral researchers, research assistants, and master's students. The 12 participants were organised into six pairs and engaged with a robot in interactions inspired by the MuMMER dataset. The MuMMER dataset was used as the basis for the interaction design because it contained the largest number of samples among the datasets included in the offline evaluation. As explained in Section~\ref{online_taks}, in these interactions, participants were instructed to imagine themselves in a shopping mall, where the robot engaged with them by providing directions, entertaining them with quizzes, and making small talk.

We selected the best-performing model from the offline evaluation, namely the Bayesian flat neural network. In addition, we investigated whether the best-performing Bayesian hierarchical neural network (addressee-first formulation) could be perceived similarly to the flat model during interaction, beyond offline performance differences. Each interaction lasted approximately three minutes, comparable to the longest conversations in the MuMMER dataset.

Participants interacted with both the flat and hierarchical models in a randomised, counterbalanced order.
Following the interactions, participants took part in semi-structured interviews to collect qualitative feedback on the interaction experience. They also completed a questionnaire consisting of six self-designed items assessing perceived usefulness, appropriateness of the agent’s behaviour, effectiveness in managing conversational flow, the agent’s contribution to the interaction, and its ability to balance participation within the group -- the questionnaire is provided in~\ref{quest_app_2}. We opted for a self-designed scale to capture aspects specific to robotic group interaction management that existing scales do not address. The questionnaire was administered separately for each of the two interaction conditions. To evaluate differences between the two models, we analysed each questionnaire item using the paired Wilcoxon signed-rank test \cite{woolson2007wilcoxon}.


\subsection{In-person Study Results}
\begin{figure}[t!]
    \centering
    \includegraphics[
        width=0.5\textwidth,
    ]{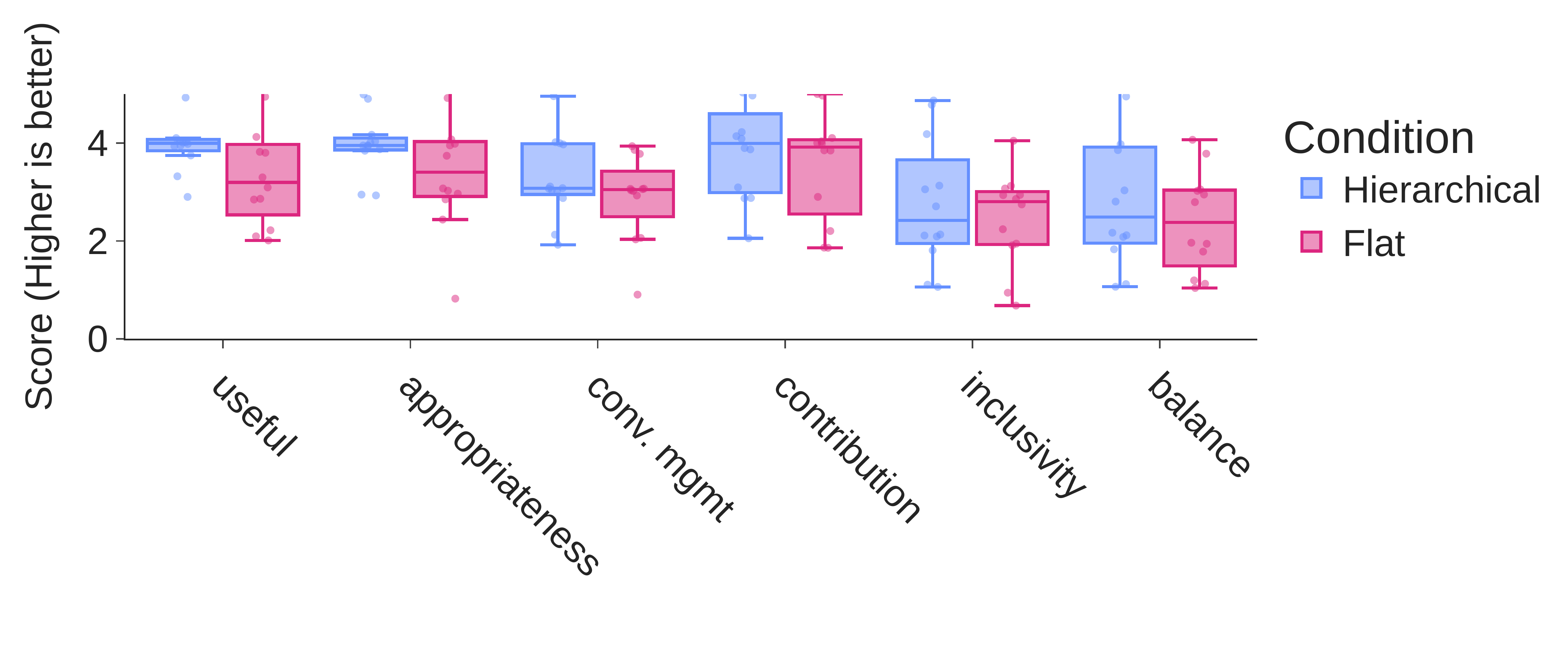}
    \caption{Box plot showing questionnaire scores for each item for the hierarchical and flat model during the in-person study.}
    \label{in person results}
\end{figure}
The results from the in-person study are shown in Figure~\ref{in person results}. Overall, participants perceived both models favourably, with most questionnaire items scoring above the midpoint of the scale (3 out of 5), and the hierarchical model consistently outperforming the flat model, albeit by moderate margins.

Appropriateness ratings were highest for the hierarchical model (M = 4.00, SD = 0.60) relative to the flat model (M = 3.42, SD = 1.16). Agent contribution followed a similar pattern, with the hierarchical model again rated somewhat higher (M = 3.83, SD = 0.94) than the flat model (M = 3.58, SD = 1.08). Perceived usefulness showed the same trend, favouring the hierarchical model (M = 4.00, SD = 0.60) over the flat model (M = 3.33, SD = 1.07).

Ratings for conversation management were more moderate for both models (hierarchical: M = 3.25, SD = 0.87; flat: M = 2.92, SD = 0.90), and conversation balancing and inclusivity received the lowest scores overall. Still, the hierarchical model held a small but consistent edge, with means of 2.83 (SD = 1.40) versus 2.42 (SD = 1.08) for balance, and 2.75 (SD = 1.36) versus 2.50 (SD = 0.90) for inclusivity.

Despite this consistent advantage for the hierarchical addressee-first Bayesian model, Wilcoxon signed-rank tests found no statistically significant differences between the two models across any of the six questionnaire dimensions (all \(p > .05\)).
Taken together, these results suggest that both models were perceived by participants as satisfactory and that the robot was able to interact effectively with participants regardless of which model was driving its behavior.

\section{Summary and Discussion}

Our results, taken together, point to complementary roles for the Bayesian and hierarchical components of HiBRIDGE. The Bayesian formulation drives \textbf{predictive accuracy}, as shown in the offline evaluation. The hierarchical formulation, by contrast, drives \textbf{interpretability}, as reflected in the online study. The in-person study extended these findings further, showing that both formulations together support 
\textbf{real-time group interaction}.

The benefit of \textbf{Bayesian modelling} is particularly relevant to HRI, where interaction data are often costly to collect and consequently limited~\cite{jospin2022hands,wilson2020bayesian}. Group interaction introduces an additional challenge because several behaviours can be reasonable in the same conversational state, making the mapping from interaction context to robot behaviour inherently difficult to learn from sparse observations. 
Consistent with this, the advantage of Bayesian modelling is strongest on the more demanding Hit@1 and Hit@3 settings and in the MuMMER~\cite{canevet2020mummer} and Spitale~\cite{spitale2026understanding} datasets, and weaker under the more lenient Hit@5 metric and in the comparatively constrained Addlesee setting~\cite{addlesee2024multi}. 
This suggests that Bayesian approaches may be particularly useful when HRI systems must learn from limited data while operating in heterogeneous or ambiguous social situations, such as group moderation~\cite{birmingham2020can}, educational settings~\cite{woo2021use}, or open-ended social interaction~\cite{abbo2025fastmultipartyopenendedconversation}. 
Although our experiments do not evaluate uncertainty calibration directly, they indicate that probabilistic parameter modelling can provide a useful \textbf{predictive advantage} in sparse HRI datasets.

The contributions of the \textbf{hierarchical formulation} instead emerge more clearly in how robot decisions can be explained. Prior work suggests that exposing intermediate computations through modular structures~\cite{swamy2023multimodn,hu2018explainable} and organising explanations hierarchically~\cite{chen2020generating} can facilitate interpretation. 
Our findings support this in HRI, with participants preferring hierarchical explanations over flat ones.
We reason that the hierarchy provides a \textbf{structured substrate for explanation} by exposing semantically meaningful intermediate decisions, such as whether the robot should act initiatively or responsively and whom it should address.
For HRI, this suggests that \textbf{interpretability} can be considered when designing the internal structure of a behaviour-selection model, rather than being solely addressed as a post-hoc process~\cite{afroogh2026beyond}. The in-person study further demonstrates that this structured prediction framework can be deployed without preventing \textbf{real-time interaction}: both the hierarchical and flat Bayesian variants were positively perceived when combined with an LLM for utterance generation. This supports a \textbf{modular approach} in which a learned model determines the robot's conversational behaviour while an LLM is used for language generation, retaining explicit control over behavioural decisions while benefiting from the flexibility of generative models.

At the same time, the interpretability benefits of hierarchy are not uniform across all measures. Overall Explanation Satisfaction Scale scores and aggregate ratings of hierarchical and flat explanations do not differ significantly, and neither explanation type improves participants' accuracy in predicting robot behaviour. Similarly, the in-person study reveals no significant differences between the hierarchical and flat Bayesian models across the evaluated interaction dimensions. One interpretation is that explanations can improve subjective understanding and preference without immediately producing a more accurate predictive mental model of the robot. The short exposure to each model may also have limited participants' opportunity to learn these behavioural patterns. Future work should therefore examine explanations over longer and repeated interactions, where participants have greater opportunity to form and refine a mental model of the robot's behaviour. This should also be studied in settings where the robot's intermediate social decisions are more consequential, such as group moderation or participation balancing ~\cite{weisswange2026design}, as the value of exposing the decision structure may become more apparent when those choices directly shape the interaction and thus explanations are more valuable to the users ~\cite{papagni2020understandable}. Moreover, because several behaviours can be plausible in the same conversational state, future work could investigate contrastive explanations~\cite{jacovi2021contrastive} that clarify why the robot selected one behaviour rather than another plausible alternative. Finally, the absence of consistent effects across all interpretability measures suggests that explanation needs may differ across users and over time, motivating adaptive explanations for different users' roles and prior experiences~\cite{10.1145/3776734.3794453}.

\section{Conclusions}

In this work, we introduced HiBRIDGE, a hierarchical Bayesian neural network framework for deciding \textit{what to say} and \textit{whom to address} in multi-party human--robot interaction. Across three offline datasets, Bayesian formulations improve predictive performance over their deterministic counterparts and outperform several baselines. While hierarchical modelling does not yield a significant predictive advantage, it exposes intermediate decision stages that provide a structured basis for explanation. Our online study shows that explanations derived from this structure are preferred over flat alternatives and perceived to be more helpful for understanding robot behaviour. The in-person study further demonstrates that HiBRIDGE can be integrated into an autonomous real-time group interaction system, with both hierarchical and flat Bayesian variants positively perceived. Taken together, these findings show that probabilistic and hierarchical modelling contribute complementary benefits: the former strengthens behaviour prediction under sparse interaction data, while the latter supports greater transparency in how those behaviours are selected. HiBRIDGE therefore provides a foundation for group-robot dialogue management that combines predictive performance, structured decision-making, and interpretable robot behaviour.

\section*{Acknowledgements}
\textbf{Contributions:} Conceptualisation: FID, HG, MS. Methodology: MN, FID, HG, MS. Software, Validation, Data Curation, Formal analysis \& Investigation: MN. Writing – original draft: MN, FID. Visualisation: MN, FID. Writing – review \& editing: MN, FID, HG, MS. Supervision \& Project administration: MS, FID, HG. Funding acquisition: MS, HG.
\textbf{Funding:} MN and MS have been funded by the PNRR-PE-AI FAIR project, within NextGeneration EU program. FID and HG are supported by CHANSE and NORFACE through the MICRO project, funded by ESRC/UKRI grant ref.~UKRI572. 
\textbf{Open access:} For the purpose of open access, the authors have applied a Creative Commons Attribution (CC BY) license to any Accepted Manuscript version arising. 
\textbf{Data access:} The datasets used for the offline evaluation are either publicly available or available upon request, depending on the dataset. Data collected through the online and in-person empirical studies are available from the authors upon reasonable request.

\section*{Declaration of Generative AI and AI-assisted Technologies in the Manuscript Preparation Process} During the preparation of this work, the authors used Claude and ChatGPT in order to assist with rephrasing and grammar correction during the writing phase. No new content was generated by the AI tools. After using these tools, the authors reviewed and edited the content as needed and take full responsibility for the content of the published article.

\bibliographystyle{elsarticle-num}   
\bibliography{bibliography}  
\clearpage

\appendix

\section{Feature Representation}
\label{app:feature_set}
 
This appendix details the full feature set used to construct the conversational representation $\mathbf{x}_t$ described in Section ~\ref{sec:feature_representation}. Table~\ref{tab:feature_set_table} reports all per-turn and interaction-level features and how they combine to reach the final dimensionality of 158. Table~\ref{tab:feature_categories} lists, for each categorical feature, the category values used during one-hot encoding.
 
\begin{table*}[h!]
\centering
\caption{Feature set composition (158 features total).}
\label{tab:feature_set_table}
\begin{tabular}{lccc}
\toprule
\textbf{Feature} & \textbf{Type} & \textbf{Width} & \textbf{Total} \\
\midrule
\multicolumn{4}{l}{\textit{Per-turn features (repeated for each of the 6 most recent turns)}} \\
\midrule
Speaker         & Categorical (4 categories)  & 4  & 24 \\
Addressee        & Categorical (5 categories)  & 5  & 30 \\
Dialogue act      & Categorical (11 categories) & 11 & 66 \\
Turn present & Binary                     & 1  & 6  \\
Utterance duration & Numerical                  & 1  & 6  \\
Silence duration   & Numerical                  & 1  & 6  \\
\cmidrule{4-4}
 & & \textit{Subtotal} & \textit{138} \\
\midrule
\multicolumn{4}{l}{\textit{Interaction-level features (computed once, over the conversation so far)}} \\
\midrule
Time spoken between each pair of members & Numerical & 12 & 12 \\
Time each member was addressed (in-degree) & Numerical & 4  & 4  \\
Time each member spoke (out-degree)        & Numerical & 4  & 4  \\
\cmidrule{4-4}
 & & \textit{Subtotal} & \textit{20} \\
\midrule
\textbf{Total} & & & \textbf{158} \\
\bottomrule
\end{tabular}
\end{table*}
 
\begin{table*}[h!]
\centering
\caption{Category values for each categorical feature.}
\label{tab:feature_categories}
\begin{tabular}{p{3.2cm}p{3.2cm}p{5.5cm}}
\toprule
\textbf{Speaker (4)} & \textbf{Addressee (5)} & \textbf{Dialogue act (11)} \\
\midrule
Robot & Robot & Statement \\
Participant 1 & Participant 1 & Factual Question \\
Participant 2 & Participant 2 & Opinion Question \\
Participant 3 & Participant 3 & Offer \\
 & \textsc{All} & Small Talk \\
 & & Opinion \\
 & & Comment \\
 & & Follow Up \\
 & & Positive Answer \\
 & & Negative Answer \\
 & & Other Answer \\
\bottomrule
\end{tabular}
\end{table*}

\section{Prompts for generating explanations}
\label{prompt_app}

This appendix documents the structure of the \textbf{system prompt} and \textbf{user prompt} used to generate natural-language explanations of a conversational robot's addressee and utterance-type decisions during a 3-party interaction (robot + left participant + right participant).

\subsection{System Prompt}

The system prompt is a fixed template with one variable, \texttt{n\_words} (default \texttt{150}), controlling the target length of the generated explanation.

{\scriptsize
\begin{verbatim}
You explain a conversational robot's
decisions during a 3-party interaction
(robot + two participants: left and right).

CONTEXT
The robot must decide:
1) Who to address (left, right, or both)
2) What kind of utterance to produce

You will receive:
- The robot's final decision
- One or more decision paths from decision
  trees
- The sequence of recent speakers

Refer to participants only as "left" or
"right".

TASK
Write a clear, concise explanation of the
robot's decision.

FORMAT (MANDATORY -- must follow exactly)
How the robot decided who to address?
The robot decided ... because ...

How the robot decided what to say?
The robot decided ... because ...

RULES
- Use exactly the two subtitles above (no
  changes)
- Keep this exact order
- Do not add extra sections or text before,
  between, or after them

REQUIREMENTS
- Base the explanation on the decision paths
  (features and thresholds), but keep it
  understandable for a general audience
- Include only the most relevant features
  (maximum of 2) to improve clarity
- Use simple, non-technical language
- Exactly {n_words} words, divided equally
  between "How the robot decided who to
  address?" and "How the robot decided what
  to say?"

STYLE CONSTRAINTS
- Do NOT use model labels (e.g., SPEAK,
  INITIATIVE, RESPONSIVE, etc.)
- Use natural phrasing:
- "address the left/right participant" or
  "address both participants"
- "start a new topic" or "continue the
  current topic"
- "ask a question", "make a statement",
  "respond", etc.

REFERENCES:

MODEL HIERARCHY
The model can be structured either as 4
sequential decision trees:
1. SPEAK -- should the robot address one
   person (INDIVIDUAL) or everyone (ALL)?
2. ALL-INDIVIDUAL -- Should the robot
   initiate a new topic (INITIATIVE) or
   respond to an ongoing one (RESPONSIVE)?
3. INITIATIVE -- if initiating: ask a
   Question or make a Statement?
4. RESPONSIVE -- if responding: give an
   Acknowledgment, Answer, or Follow Up?

or as one decision tree deciding for who to
address and what to say at the same time.

DIALOGUE ACTS
- Statement: start a topic or social talk
- Follow Up: continue the current topic with
  a question
- Question: ask for information, opinions,
  small talk, or make an offer
- Answer: respond to a question
- Acknowledgment: react to prior context
  without changing topic

ADDRESSEE TYPES
- IND: one specific person
- ALL: everyone in the group

FEATURE GUIDE
The model uses 47 features from the last 5
turns and global conversation state.

Participants: 0=ROBOT, 1=part_1, 2=part_2,
3=part_3

SEQUENTIAL FEATURES (one-hot, index 0=5
turns ago, index 4=last turn)
- Speaker:
  seq_chunk_speaker_{turn}_{participant}
- Addressee:
  seq_chunk_addressee_{turn}_{participant/ALL}
- Dialogue act:
  seq_chunk_utterance_type_{turn}_{act}

LAST TURN FEATURES (one-hot, repeat of most
recent turn)
- token_last_speaker_{participant}
- token_last_addressee_{participant/ALL}
- token_last_utterance_type_{act}

GLOBAL FEATURES (numerical, mean-std
normalized, 0=ROBOT)
- global_out_degree_{i} -> total speech by
  participant i
- global_in_degree_{i} -> total speech
  directed at participant i
- global_time_edge_{i}_{j} -> how much
  participant i has spoken to participant j
\end{verbatim}
}

\subsection{User Prompt}
The user prompt has a fixed structure:

{\scriptsize
\begin{verbatim}
This is the model explanation:
{decision_tree_path} and this is the
speaker list: {speaker_list}
\end{verbatim}
}

Where \texttt{\{decision\_tree\_path\}} is the path taken during the prediction by the surrogate decision tree model(s), either \textbf{hierarchical} (multiple sequential decision paths) or \textbf{flat} (a single decision path), and \texttt{\{speaker\_list\}} is the sequence of recent speakers.




\section{Explanation Generation Examples}
\label{app:explanation-examples}
 
This appendix shows two worked examples of the explanation-generation pipeline: the decision-tree path(s) that form the \texttt{\{decision\_tree\_path\}} input to the system, and the resulting natural-language explanation produced in the two-part format.
 
\subsection{Example 1: Hierarchical Decision Path}
 
\subsubsection*{Decision-Tree Path (Input Example)}
 
{\scriptsize
\begin{verbatim}
This is the model explanation: Decision path
for head 'SPEAK':
Predicted class: ALL
----------------------------------------
Node 0: seq_chunk_speaker_3_ROBOT = 0.000 <=
    0.500 -> go left
Node 1: seq_chunk_addressee_2_ALL = 1.000 >
    0.500 -> go right
Node 35: seq_chunk_utterance_type_4_Follow
    Up = 0.000 <= 0.500 -> go left
Node 36: seq_chunk_utterance_type_3_Comment
    = 0.000 <= 0.500 -> go left
Node 37: global_in_degree_2 = 1.163 > 1.065
    -> go right
=> Leaf node 45: predict class ALL
 
Decision path for head 'ALL-INDIVIDUAL':
Predicted class: RESPONSIVE
----------------------------------------
Node 0: seq_chunk_addressee_4_ROBOT = 0.000
    <= 0.500 -> go left
Node 1: global_in_degree_0 = 2.014 > -0.904
    -> go right
Node 5: seq_chunk_addressee_1_part_3 = 0.000
    <= 0.500 -> go left
Node 6: seq_chunk_utterance_type_2_Follow Up
    = 0.000 <= 0.500 -> go left
Node 7: seq_chunk_utterance_type_4_Comment =
    0.000 <= 0.500 -> go left
Node 8: seq_chunk_addressee_1_ALL = 0.000 <=
    0.500 -> go left
Node 9: seq_chunk_speaker_3_part_2 = 1.000 >
    0.500 -> go right
Node 15: global_out_degree_2 = 0.595 >
    -0.009 -> go right
=> Leaf node 17: predict class RESPONSIVE
 
Decision path for head 'RESPONSIVE':
Predicted class: FOLLOW UP
----------------------------------------
Node 0: seq_chunk_utterance_type_4_Follow Up
    = 0.000 <= 0.500 -> go left
Node 1: seq_chunk_utterance_type_4_Factual
    Question = 0.000 <= 0.500 -> go left
Node 2: seq_chunk_addressee_4_ROBOT = 0.000
    <= 0.500 -> go left
Node 3: seq_chunk_duration_4 = 0.276 <=
    0.302 -> go left
Node 4: global_time_edge_0_2 = -0.549 <=
    0.093 -> go left
Node 5: global_out_degree_1 = 0.002 > -0.451
    -> go right
=> Leaf node 7: predict class FOLLOW UP and
    this is the speaker list: ['left',
    'right', 'robot']
\end{verbatim}
}
 
\subsubsection*{Generated Explanation (Output Example)}
 
\begin{quote}
\textbf{How did the robot decide who to address?}
 
The robot decided to address both participants together. It had already spoken to both of them at once a little earlier in the conversation, and the right participant had been receiving noticeably more attention from the group than usual. Addressing both together kept the discussion consistent and made sure that attention stayed shared rather than going to just one person.
 
\textbf{How did the robot decide what to say?}
 
The robot decided to continue the current topic by asking a follow-up question rather than starting something new. It had recently been the focus of a lot of the group's attention, and the right participant had just spoken and had generally been quite active. Since the discussion was clearly still ongoing, following up with a question kept both participants engaged.
\end{quote}
 
\subsection{Example 2: Flat Decision Path}
 
\subsubsection*{Decision-Tree Path (Input Example)}
 
{\scriptsize
\begin{verbatim}
This is the model explanation: Predicted
class: FOLLOW UP/ALL
----------------------------------------
Node 0: seq_chunk_duration_1 = 0.133 >
    -2.528 -> go right
Node 2: seq_chunk_utterance_type_4_Follow Up
    = 0.000 <= 0.500 -> go left
Node 3: seq_chunk_addressee_4_ROBOT = 0.000
    <= 0.500 -> go left
Node 4: global_in_degree_1 = 0.475 > -1.000
    -> go right
Node 10: seq_chunk_addressee_4_ALL = 1.000 >
    0.500 -> go right
Node 30: seq_chunk_utterance_type_4_Comment
    = 0.000 <= 0.500 -> go left
Node 31: seq_chunk_duration_1 = 0.133 <=
    0.178 -> go left
=> Leaf node 32: predict class FOLLOW UP/ALL
    and this is the speaker list: ['left',
    'right', 'robot']
\end{verbatim}
}
 
\subsubsection*{Generated Explanation (Output Example)}
 
\begin{quote}
\textbf{How did the robot decide who to address?}
 
The robot decided to address both participants together again. Its most recent turn had also been directed at the whole group rather than a single person, keeping that same pattern. The left participant had also been receiving a reasonable amount of attention from others earlier on, so continuing to speak to everyone kept the conversation balanced.
 
\textbf{How did the robot decide what to say?}
 
The robot decided to continue the current topic by asking a follow-up question rather than starting something new or making a comment. The conversation had been unfolding for a while without becoming repetitive, and the most recent turn hadn't itself been a follow-up or a passing comment, so asking a further question kept the discussion moving forward.
\end{quote}

\section{Hyperparameters optimised during experiments}
\label{param_app}

\begin{table*}[ht!]
\centering
\small
\caption{Best hyperparameters for the deterministic models.}
\label{tab:hparams-det}
\begin{tabular}{lcccc}
\hline
\textbf{Hyperparameter} & \textbf{Speak/No-Speak} & \textbf{Flat} & \textbf{Addressee-First} & \textbf{Dialogue-Act-First} \\
\hline
Number of hidden layers            & 1    & 1    & 1      & 1      \\
Number of neurons per layer        & 75   & 25   & 50     & 50     \\
Dropout rate                       & 0.2  & 0.6  & 0.4    & 0.4    \\
L2 regularisation strength         & 0.1  & 0.1  & 0.0001 & 0.0001 \\
Class-weighted cross-entropy       & ---  & False& False  & False  \\
Cross-entropy weight parameter     & ---  & 0.5  & 0.5    & 0.5    \\
Max class weight                   & ---  & 2.0  & 2.0    & 2.0    \\
Number of augmented data copies    & 3    & 3    & 3      & 3      \\
Magnitude of injected noise        & 0.05 & 0.05 & 0.02   & 0.02   \\
Hierarchical loss penalty coeff.   & ---  & 0    & 0      & 0      \\
\hline
\end{tabular}
\end{table*}

\begin{table*}[ht!]
\centering
\small
\caption{Best hyperparameters for the Bayesian models.}
\label{tab:hparams-bayes}
\begin{tabular}{lccc}
\hline
\textbf{Hyperparameter} & \textbf{Flat} & \textbf{Addressee-First} & \textbf{Dialogue-Act-First} \\
\hline
Number of hidden layers            & 1    & 2      & 2      \\
Number of neurons per layer        & 50   & 75     & 25     \\
Dropout rate                       & 0.6  & 0.2    & 0.4    \\
Learning rate                      & 0.001& 0.0001 & 0.0002 \\
Prior std.\ dev.\ --- weights      & 4.0  & 3.0    & 3.0    \\
Prior std.\ dev.\ --- bias         & 2.0  & 5.0    & 2.0    \\
Number of augmented data copies    & 2    & 3      & 2      \\
Magnitude of injected noise        & 0.04 & 0.04   & 0.04   \\
Cross-entropy component weighting  & 1.0  & 0.5    & 1.0    \\
Hierarchical loss penalty coeff.   & 4    & 8      & 8      \\
\hline
\end{tabular}
\end{table*}

\begin{table*}[ht!]
\centering
\small
\caption{Best hyperparameters for the Gillet et al.\ baseline.}
\label{tab:hparams-gillet}
\begin{tabular}{lccc}
\hline
\textbf{Hyperparameter} & \textbf{Shared} & \textbf{Addressee} & \textbf{Utterance-type} \\
\hline
Loss function                      & Cross-entropy & --- & --- \\
Batch size                         & 64            & --- & --- \\
Use decision threshold             & False         & --- & --- \\
Data augmentation                  & True          & --- & --- \\
Number of dataset replications     & 2             & --- & --- \\
Amount of noise injected           & 0.4           & --- & --- \\
First-layer message architecture   & ---           & 8-16   & 8-16   \\
First-layer node architecture      & ---           & 8-16-4 & 8-16-4 \\
Second layer used                  & ---           & True   & False  \\
Second-layer message architecture  & ---           & 8-16   & 8-16   \\
Second-layer node architecture     & ---           & 4-2    & 4-2    \\
Dropout rate                       & ---           & 0.5    & 0.2    \\
Learning rate                      & ---           & 0.0004 & 0.0008 \\
\hline
\end{tabular}
\end{table*}

This appendix reports the best hyperparameter configuration found for each model, as selected by Optuna~\cite{akiba2019optuna} over 75 trials, using average validation accuracy from 10-fold cross-validation on the MuMMER dataset (see Section~\ref{imp_details}, \emph{Hyperparameter Optimisation}). We report the hyperparameters listed in that subsection for each model family. Entries marked ``---'' indicate a hyperparameter that does not apply to that model.

\textbf{Deterministic Models:} 
Table~\ref{tab:hparams-det} reports the best configuration for the Speak/No-Speak gate, deterministic Flat model and the two deterministic hierarchical architectures (Addressee-First, Dialogue-Act-First).

\textbf{Bayesian Hierarchical Models:} Table~\ref{tab:hparams-bayes} reports the best configuration for the three Bayesian architectures: Flat, Addressee-First, and Dialogue-Act-First, respectively.

\textbf{Gillet et al.\ Baseline:} This subsection reports the best hyperparameter configuration found for the Gillet et al.\ baseline~\cite{gillet2025templates}, as selected by Optuna~\cite{akiba2019optuna} using average validation accuracy from 10-fold cross-validation on the MuMMER dataset (see Section~\ref{imp_details}, \emph{Hyperparameter Optimisation}). Table~\ref{tab:hparams-gillet} reports the hyperparameters shared across both sub-models together with those tuned separately for the addressee and utterance-type sub-models.

\section{Custom-Designed Online Study Questionnaire}
\label{ques_app}

Participants rated the following statements on a 5-point Likert scale, ranging from 1 (Strongly Disagree) to 5 (Strongly Agree) for the online interpretability study.

\texttt{Please rate the following statements about the explanation:}

\begin{enumerate}
    \item [\texttt{1)}]\texttt{Explanations were clear and understandable.}

    \item [\texttt{2)}]\texttt{Explanations helped me to understand how the robot chose who to address.}

    \item [\texttt{3)}]\texttt{Explanations helped me to understand how the robot chose what to say.}

    \item [\texttt{4)}]\texttt{I can now understand the robot's behaviour better.}

    \item [\texttt{5)}]\texttt{The explanation was coherent with the robot's decision on who to address.}

    \item [\texttt{6)}]\texttt{The explanation was coherent with the robot's decision on what to say.}
\end{enumerate}

\newlength{\shotheight}
\setlength{\shotheight}{9cm}

\begin{figure*}[t!]
\centering
\begin{minipage}{0.22\linewidth}
\centering
\parbox[b][\shotheight][c]{\linewidth}{\centering\includegraphics[width=\linewidth]{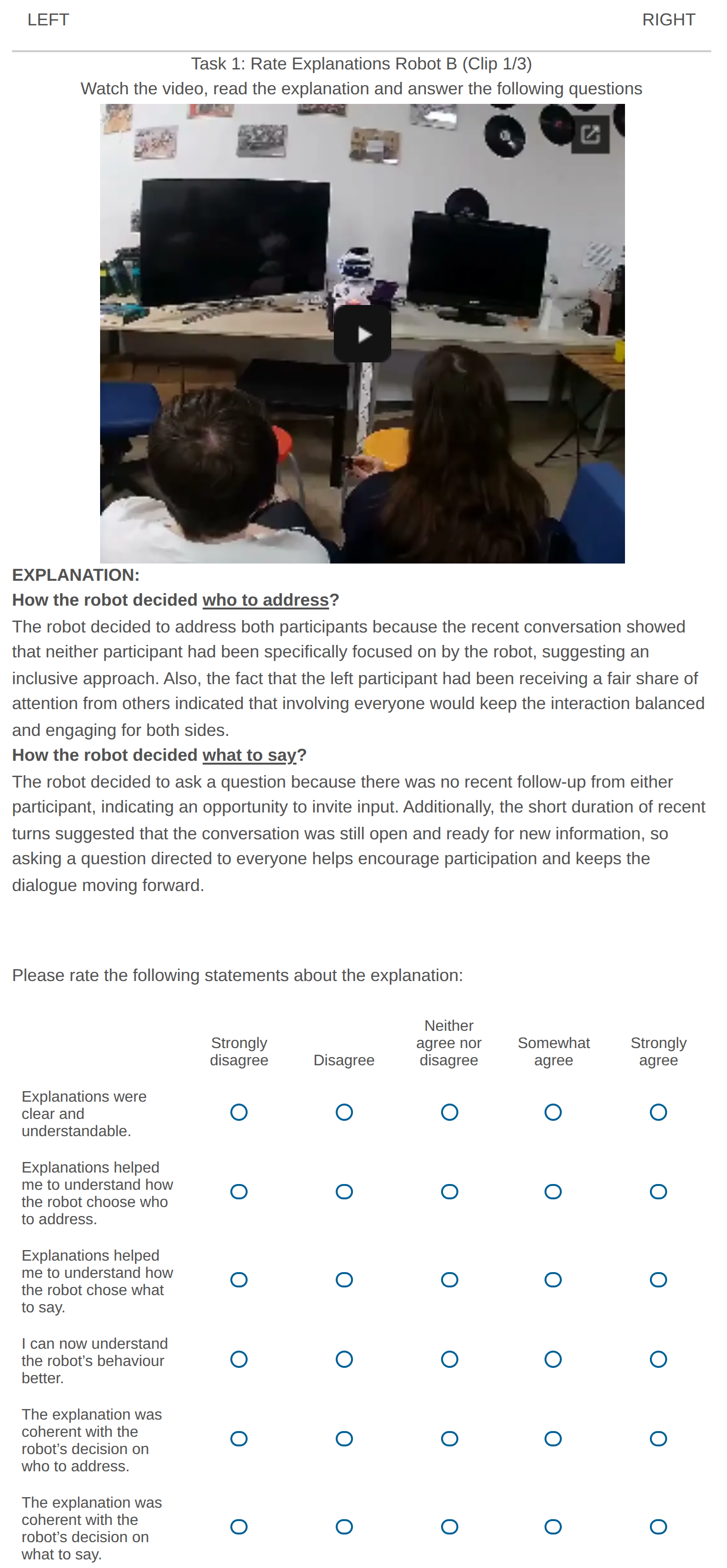}}
\caption*{(a) Task 1}
\end{minipage}\hfill
\begin{minipage}{0.21\linewidth}
\centering
\parbox[b][\shotheight][c]{\linewidth}{\centering\includegraphics[width=\linewidth]{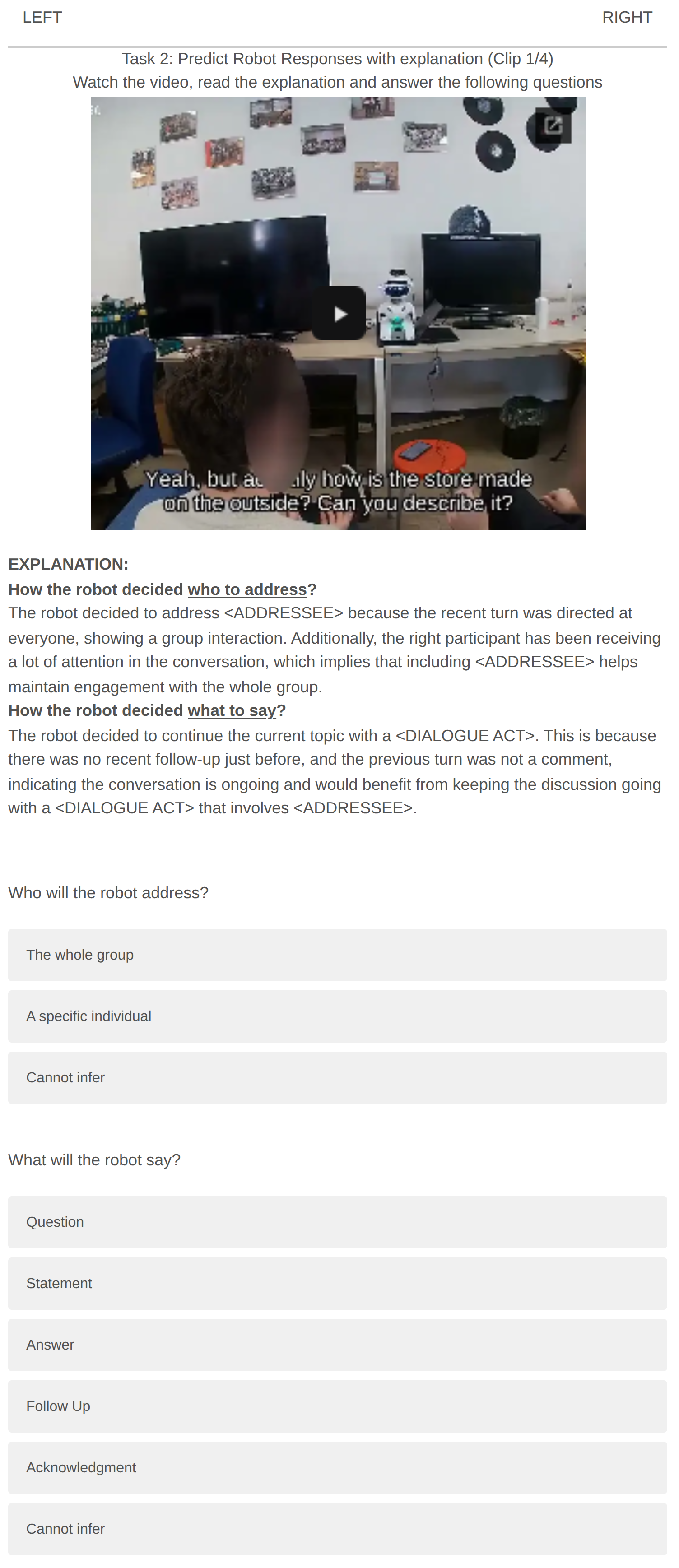}}
\caption*{(b) Task 2}
\end{minipage}\hfill
\begin{minipage}{0.26\linewidth}
\centering
\parbox[b][\shotheight][c]{\linewidth}{\centering\includegraphics[width=\linewidth]{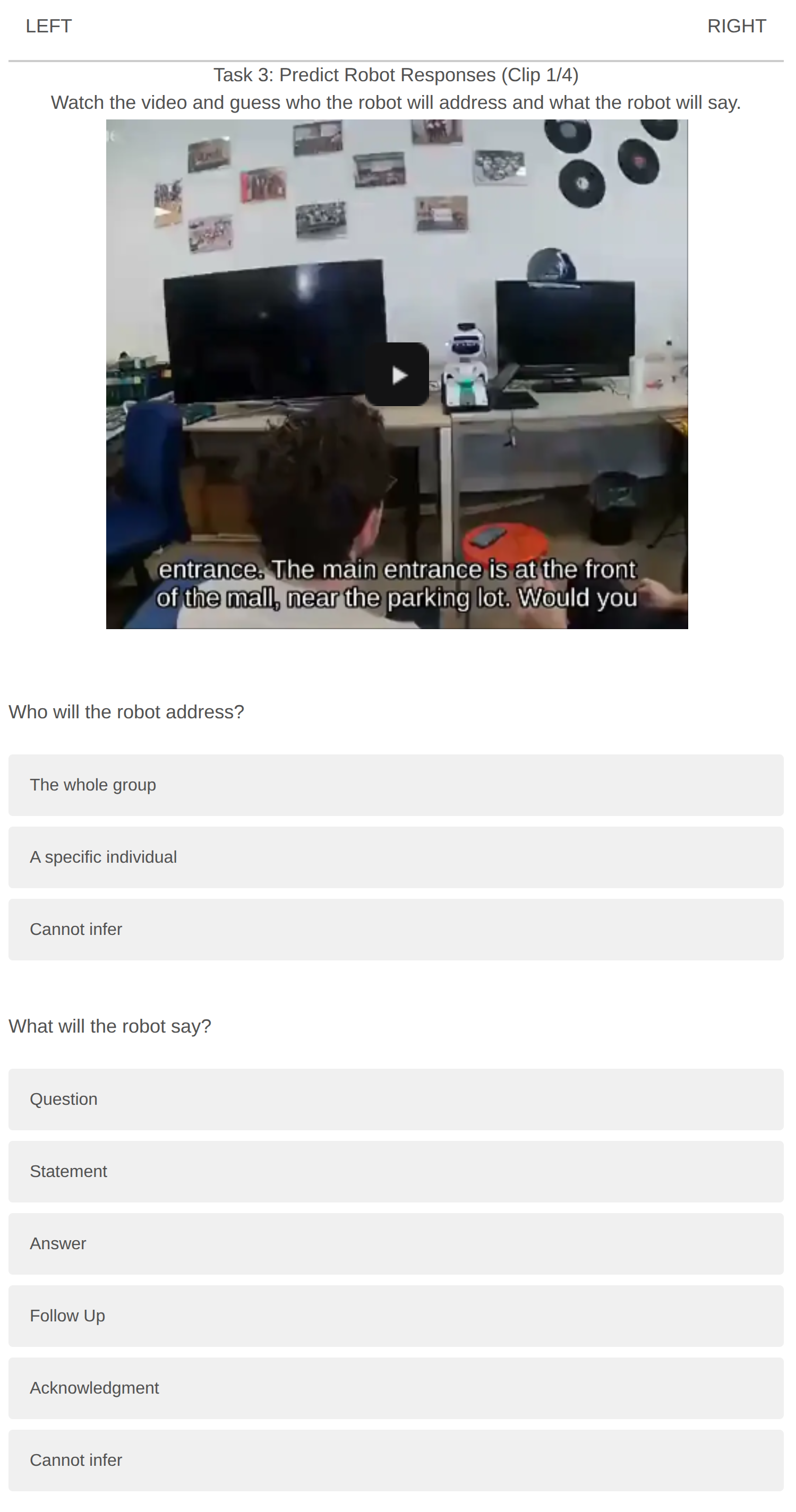}}
\caption*{(c) Task 3 (part 1)}
\end{minipage}\hfill
\begin{minipage}{0.28\linewidth}
\centering
\parbox[b][\shotheight][c]{\linewidth}{\centering\includegraphics[width=\linewidth]{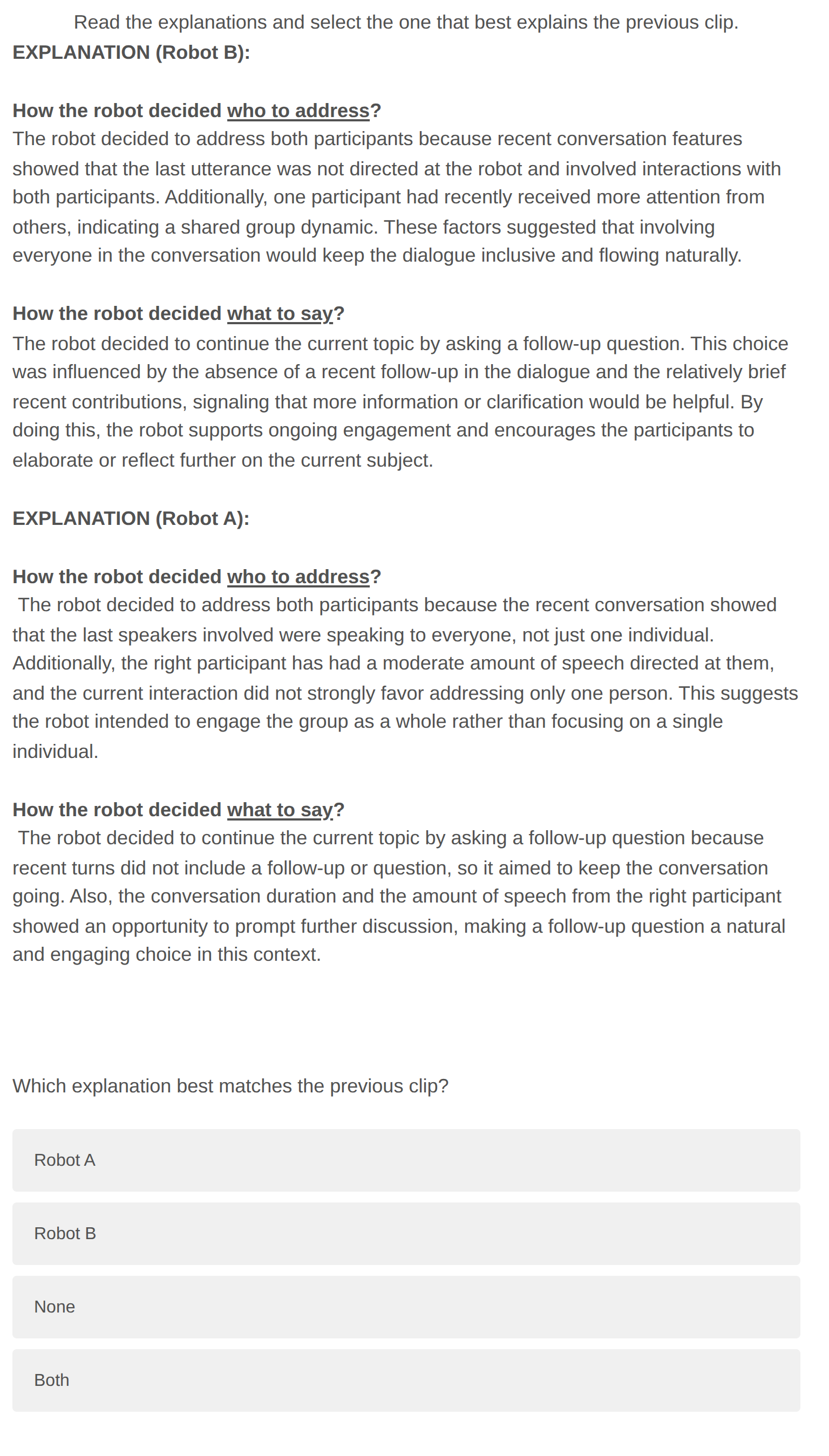}}
\caption*{(d) Task 3 (part 2)}
\end{minipage}
\caption{Example screenshots of the Qualtrics questionnaire interface: (a) explanation-rating questions in Task 1; (b) the behaviour-prediction question in Task 2; (c) the behaviour-prediction question in Task 3; (d) the side-by-side explanation preference question in Task 3.}
\label{fig:qualtrics_screenshots}
\end{figure*}
\section{Online Study Task Interface}
\label{app:qualtrics_screenshots}
The online interpretability study was implemented and hosted on Qualtrics. Figure~\ref{fig:qualtrics_screenshots} shows example screenshots of the interface presented to participants for each of the three tasks.

\section{Custom-Designed In-Person Study Questionnaire}
\label{quest_app_2}

Participants rated the following items on a 5-point Likert scale after each part of the in-person study.
 
\begin{enumerate}
    \item [\texttt{1)}]\texttt{How useful did you find the robot for this specific application?} \\
    \texttt{(1 = Not Useful at all, 5 = Highly Useful)}
 
    \item [\texttt{2)}]\texttt{How appropriate was the agent's behavior within the group conversation?} \\
    \texttt{(1 = Not Appropriate, 5 = Highly Appropriate)}
 
    \item [\texttt{3)}]\texttt{How effectively did the agent manage the flow and dynamics of the group conversation?} \\
    \texttt{(1 = Poorly Managed, 5 = Effectively Managed)}
 
    \item [\texttt{4)}]\texttt{How did you perceive the level of the agent's engagement and contribution to the conversation?} \\
    \texttt{(1 = Not Active, 5 = Highly Active)}
 
    \item [\texttt{5)}]\texttt{How effectively did the robot engage and include all participants in the conversation?} \\
    \texttt{(1 = Not Effectively, 5 = Very Effectively)}
 
    \item [\texttt{6)}]\texttt{How equally did the robot balance participation among all participants during the interaction?} \\
    \texttt{(1 = Not Equally, 5 = Very Equally)}
\end{enumerate}


\end{document}